\documentclass[letterpaper, 10 pt, conference]{ieeeconf}  

\usepackage{cite}
\usepackage{bm}
\usepackage{makecell}
\usepackage{amsmath} 
\usepackage{amssymb}  
\usepackage{amsfonts}
\usepackage{ multicol} 

\usepackage{graphics} 
\usepackage{epsfig} 
\usepackage{mathptmx} 
\usepackage{times} 

\usepackage{mathrsfs}
\usepackage{indentfirst}
\usepackage{multirow}
\usepackage{adjustbox}

\usepackage{color}      
\usepackage{pifont}     
\usepackage{algorithmic}
\usepackage{graphicx}
\usepackage{textcomp}
\usepackage{bbding}

\usepackage[table,dvipsnames,xcdraw]{xcolor} 

\usepackage{colortbl} 
\usepackage{diagbox} 
\usepackage{siunitx}

\usepackage[colorlinks=true,linkcolor=blue,citecolor=green,urlcolor=blue,]{hyperref}

\usepackage{algorithm}
\usepackage{algorithmic}

\usepackage{subcaption}

\colorlet{colorFst}{Green!25}       
\colorlet{colorSnd}{SpringGreen!45} 
\colorlet{colorTrd}{Yellow!30}      
\colorlet{colorLow}{darkgray!30}    
\definecolor{R1}{HTML}{E97451}
\definecolor{R2}{HTML}{008080}
\definecolor{R3}{HTML}{0047AB}
\colorlet{cmt}{darkgray!80}    
\colorlet{supp}{darkgray!50}    

\newcommand{\blackx}{{\color{black}\ding{55}}}

\newcommand{\fs}{\cellcolor{colorFst}}   
\newcommand{\nd}{\cellcolor{colorSnd}}      
\newcommand{\rd}{\cellcolor{colorTrd}}      

\IEEEoverridecommandlockouts                              

\title{\LARGE \bf
Towards Robust Sensor-Fusion Ground SLAM: A Comprehensive Benchmark and A Resilient Framework
}

\author{Deteng Zhang$^{1\dagger}$, Junjie Zhang$^{2\dagger}$, Yan Sun$^{3}$, Tao Li$^{4}$, Hao Yin$^{5}$, Hongzhao Xie$^{5}$ and Jie Yin$^{5}$* 
\thanks{$^*$ Corresponding author: \textbf{Jie Yin ({\tt\small \textcolor{magenta}{robot\_yinjie@outlook.com}})}}
\thanks{$^\dagger$ Equal contribution. $^{1}$Independent, $^{2}$Chongqing University, $^{3}$Nankai University, $^{4}$Zhejiang University of Technology,
$^{5}$Shanghai Jiao Tong University}%
}

\begin{document}

\maketitle
\thispagestyle{empty}
\pagestyle{empty}

\begin{abstract}

Considerable advancements have been achieved in SLAM methods tailored for structured environments, yet their robustness under challenging corner cases remains a critical limitation. Although multi-sensor fusion approaches integrating diverse sensors have shown promising performance improvements, the research community faces two key barriers: On one hand, the lack of standardized and configurable benchmarks that systematically evaluate SLAM algorithms under diverse degradation scenarios hinders comprehensive performance assessment. While on the other hand, existing SLAM frameworks primarily focus on fusing a limited set of sensor types, without effectively addressing adaptive sensor selection strategies for varying environmental conditions.

To bridge these gaps, we make three key contributions: First, we introduce \href{https://github.com/sjtuyinjie/M3DGR}{M3DGR} dataset: a sensor-rich benchmark with systematically induced degradation patterns including visual challenge, LiDAR degeneracy, wheel slippage and GNSS denial. Second, we conduct a comprehensive evaluation of forty SLAM systems on M3DGR, providing critical insights into their robustness and limitations under challenging real-world conditions. Third, we develop a resilient modular multi-sensor fusion framework named \href{https://github.com/sjtuyinjie/Ground-Fusion2}{Ground-Fusion++}, which demonstrates robust performance by coupling GNSS, RGB-D, LiDAR, IMU (Inertial Measurement Unit) and wheel odometry. Codes \textbf{}\footnote{\href{https://github.com/sjtuyinjie/Ground-Fusion2}{https://github.com/sjtuyinjie/Ground-Fusion2}} and datasets \textbf{}\footnote{\href{https://github.com/sjtuyinjie/M3DGR}{https://github.com/sjtuyinjie/M3DGR}} are publicly available.

\end{abstract}

\section{INTRODUCTION}

Ground robots demonstrate significant potential in executing complex tasks such as industrial inspection, catering services, and agricultural automation\cite{cadena2016past,al2024review,yin2023design}, where robust localization is the most fundamental requirement. Recent advances in multi-sensor fusion SLAM (Simultaneous Localization and Mapping)\cite{lin2021r3live,yin2023sky,qu2024implicit} have shown promising results in controlled environments. However, as revealed in\cite{khedekar2022mimosa,yin2024ground}, these systems exhibit critical vulnerabilities when confronting real-world challenges like temporary sensor failures or environmental extremes.

Existing benchmarking efforts face dual limitations:
From the dataset perspective, predominant datasets\cite{schubert2018tum, yin2023ground} emphasize common operational scenarios with fixed sensor configurations, lacking configurable rich sensor suites for algorithm customization and degradation scenarios for failure analysis. On the algorithmic front,  although multi-sensor fusion strategies (loosely/tightly-coupled) which leverage complementary characteristics of heterogeneous sensors are demonstrated to be effective in enhancing localization performance, the question of optimal sensor selection\cite{yin2023ground,yin2024ground} under specific degradation scenarios remains relatively understudied.

    

To fill these gaps, we introduce a novel challenging SLAM dataset {\textbf{M3DGR}: a \textbf{M}ulti-sensor, \textbf{M}ulti-scenario and \textbf{M}assive-baseline SLAM dataset for \textbf{G}round \textbf{R}obot with a full sensor suite, incorporating systematically designed corner cases that induce sensor degradation scenarios. Furthermore, we conduct extensive evaluations of cutting-edge SLAM algorithms across diverse sensor configurations based on our dataset, providing critical insights into their respective limitations. Moreover, we develop a modular multi-sensor fusion SLAM framework named \textbf{Ground-Fusion++} to facilitate future research in SLAM. We summarize the core contributions of this work as follows:

\begin{itemize}
    \item  We launch a comprehensive SLAM benchmark named M3DGR for ground robots, featuring a rich sensor suite that includes an RGBD-IMU sensor, an omnidirectional camera, two non-repetitive scanning LiDARs, a wheel odometer, and two GNSS receivers. Furthermore, M3DGR covers challenging sensor-degeneration scenarios, such as visual degradation, LiDAR-degeneracy, wheel slippage and GNSS-denial.

    \item  We conduct systematic evaluation with \textbf{forty} cutting-edge SLAM algorithms with diverse sensor configurations on our dataset. A thorough analysis is performed to evaluate these algorithms, as to provide key insights to their robustness and limitations under diverse challenging scenarios.

    \item  We develop a resilient modular SLAM framework named Ground-Fusion++, which integrates GNSS, RGB-D camera, IMU, wheel odometer and LiDAR. This system achieves robust localization and promising mapping in large-scale environments, establishing a strong and adaptable baseline for future research in multi-sensor fusion SLAM.

\end{itemize}

 \begin{table*}[ht]
    \small
    \centering
    \caption{Comparison with other SLAM benchmark datasets.}
    \label{Comparison}
    \renewcommand{\arraystretch}{1.3} 
    \begin{adjustbox}{width=1.65\columnwidth}
        \begin{tabular}{*{14}c}
            \hline
            \multirow{2}{*}{\makecell{Dataset/Year}} & \multicolumn{5}{c}{\makecell{Scenario}}  & \multicolumn{7}{c}{\makecell{Sensors}} & \multirow{2}{*}{\makecell{Number of \\ compared algorithms }} \\
                \cline{2-5} \cline{7-13}
                & VC$^1$ & LR$^2$ & GD$^3$ & WS$^4$ &  & RGB & Depth & Omni$^5$ & IMU & LiDAR & Wheel & GNSS &  \\
            \hline

            \makecell{EuRoC\cite{burri2016euroc}, 2016} & \makecell{\Checkmark} & \makecell{} & \makecell{} & \makecell{} & \makecell{} & \makecell{\Checkmark} & \makecell{} & \makecell{} & \makecell{\Checkmark} & \makecell{}  & \makecell{} & \makecell{} & \makecell{0} \\

            \makecell{UrbanLoco\cite{wen2020urbanloco}, 2020} & \makecell{\Checkmark} & \makecell{} & \makecell{\Checkmark} & \makecell{} & \makecell{ } & \makecell{\Checkmark} & \makecell{} & \makecell{} & \makecell{\Checkmark} & \makecell{1}  & \makecell{} & \makecell{\Checkmark} & \makecell{3} \\
                
            \makecell{OpenLoris-Scene\cite{shi2020we}, 2020} & \makecell{\Checkmark} & \makecell{} & \makecell{} & \makecell{\Checkmark} & \makecell{ } & \makecell{\Checkmark} & \makecell{\Checkmark} & \makecell{} & \makecell{\Checkmark} & \makecell{}  & \makecell{\Checkmark} & \makecell{} & \makecell{9} \\
            
            \makecell{M2DGR\cite{yin2021m2dgr}, 2021} & \makecell{\Checkmark} & \makecell{\Checkmark} & \makecell{\Checkmark} & \makecell{} & \makecell{ } & \makecell{\Checkmark} & \makecell{} & \makecell{\Checkmark} & \makecell{\Checkmark} & \makecell{1}  & \makecell{} & \makecell{\Checkmark} & \nd \sl{10} \\
            
            \makecell{FusionPortable\cite{jiao2022fusionportable}, 2022} & \makecell{\Checkmark} & \makecell{} & \makecell{} & \makecell{} & \makecell{ } & \makecell{\Checkmark} & \makecell{} & \makecell{} & \makecell{\Checkmark} & \makecell{1}  & \makecell{} & \makecell{\Checkmark} & \makecell{5} \\         
            \makecell{Ground-Challenge\cite{yin2023ground}, 2023} & \makecell{\Checkmark} & \makecell{\Checkmark} & \makecell{} & \makecell{\Checkmark} & \makecell{ } & \makecell{\Checkmark} & \makecell{\Checkmark} & \makecell{} & \makecell{\Checkmark} & \makecell{}  & \makecell{} & \makecell{} & \makecell{5} \\

            \makecell{M2DGR-Plus\cite{yin2024ground}, 2024} & \makecell{\Checkmark} & \makecell{} & \makecell{\Checkmark} & \makecell{\Checkmark} & \makecell{ } & \makecell{\Checkmark} & \makecell{\Checkmark} & \makecell{} & \makecell{\Checkmark} & \makecell{1}  & \makecell{\Checkmark} & \makecell{\Checkmark} & \makecell{6} \\

            \makecell{MARS-LVIG\cite{li2024mars}, 2024} & \makecell{\Checkmark} & \makecell{\Checkmark} & \makecell{} & \makecell{} & \makecell{ } & \makecell{\Checkmark} & \makecell{} & \makecell{} & \makecell{\Checkmark} & \makecell{1}  & \makecell{} & \makecell{\Checkmark} & \makecell{6} \\

            \hline
            \makecell{\textbf{M3DGR(Ours)}, 2025} & \makecell{\Checkmark} & \makecell{\Checkmark} & \makecell{\Checkmark} & \makecell{\Checkmark} & \makecell{ } & \makecell{\Checkmark} & \makecell{\Checkmark} & \makecell{\Checkmark} & \makecell{\Checkmark} & \makecell{2}  & \makecell{\Checkmark} & \makecell{\Checkmark} & {\fs \bf{40}} \\

            \hline
        \end{tabular}
    \end{adjustbox}
    \vskip 1mm
    \footnotesize{$^1$ is visual challenge, $^2$ is LiDAR degeneracy, $^3$ is GNSS denied zone, $^4$ is wheel slippage and $^5$ is omnidirectional camera.}
    \vspace{-4.5mm}
  \end{table*}

\section{Related work}
\subsection{SLAM Benchmark Datasets}

High-quality benchmark datasets serve as critical catalysts for advancing SLAM research by enabling comprehensive comparisons. Although numerous multi-sensor SLAM datasets exist, they suffer from limitations including obsolescence, insufficient data volume, and inadequate coverage of challenging scenarios. For instance, early datasets such as EuRoC\cite{burri2016euroc}, and TUM-VI\cite{burri2016euroc} cover limited sensor types and only support evaluation of vision-based methods. The OpenLORIS-Scene\cite{shi2020we} and Ground-Challenge\cite{yin2023ground} datasets further extend sensor types, but their (pseudo-)ground-truth trajectories are calculated from LiDAR-based SLAM, which can sometimes be inaccurate. These limitations have restricted existing benchmarks to narrow evaluations of limited algorithm categories. More recently, datasets such as M2DGR\cite{yin2021m2dgr} and MARS-LVIG\cite{li2024mars} have gained attention for their richer sensor suites and more diverse environments, broadening support for more SLAM categories. However, they lack systematic evaluation for extreme conditions, limiting their effectiveness in assessing robustness in real-world degradation scenarios. As summarized in Table \ref{Comparison}, most existing benchmark do not provide sufficient algorithm coverage, leaving gaps in evaluating the adaptability of multi-sensor SLAM systems under adverse conditions. 

With the rapid progress in multi-sensor fusion SLAM \cite{xu2022review, he2025ligo, yin2023sky, hua2023m2c, yin2024ground}, there is a growing demand for a standardized benchmark with comprehensive evaluations across diverse sensor configurations. Although large-scale evaluations for SLAM are inherently tedious, time-consuming and resource-intensive, they are essential for ensuring rigorous performance assessment and driving future advancements in the field. 

To address above needs, this work introduces M3DGR, a sensor-rich dataset designed specifically to evaluate the robustness of SLAM under systematically induced sensor degradation scenarios. As shown in Table~\ref{tab:degradation_comparison}, M3DGR exhibits a higher level of degradation compared to existing datasets, providing a more challenging testbed for SLAM algorithms. Additionally, we conduct a large-scale evaluation of \textbf{forty} advanced SLAM algorithms. To ensure reproducibility, we will release all our custom algorithm implementations for M3DGR upon paper acceptance.



\begin{table}[h]
\vspace{-2.5mm}
\centering
\caption{Comparison of degradation levels between M3DGR and baseline datasets across different sensor modalities.}
\begin{adjustbox}{width=0.85\columnwidth}
\begin{tabular}{l|c|c}
\hline
\textbf{Degeneration Metric} & \textbf{M3DGR (Ours)} & \textbf{Baseline}$^1$ \\ \hline
Visual (avg. keypoints/frame)~\textbf{$\downarrow$} & \textbf{117}  & 138 \\  
LiDAR (avg. LiDAR matching error)~\textbf{$\uparrow$} & \textbf{0.14}  & 0.04 \\  
Wheel (drift rate, m)~\textbf{$\uparrow$} & \textbf{10.24}  & 2.04 \\ 
GNSS (avg. valid satellites num.)~\textbf{$\downarrow$} & \textbf{4}  & 13 \\  
\hline
\end{tabular}
\end{adjustbox}
\label{tab:degradation_comparison}
\begin{flushleft}
\footnotesize{$^1$ Visual and wheel odometry baseline sequences are from~\cite{yin2023ground}, GNSS baseline from~\cite{yin2024ground}, and LiDAR baseline from~\cite{yin2021m2dgr}. Arrows indicate higher degradation levels.}
\end{flushleft}
\vspace{-5mm}
\end{table}

\begin{table*}[ht]
    \small
    \caption{Comparison with other multi-sensor fusion SLAM systems.}
    \centering
    \renewcommand{\arraystretch}{1.3} 
    \label{slam_comparison}
    \begin{adjustbox}{width=1.65\columnwidth}
    \begin{tabular}{*{10}c}
        \hline  
        \makecell{{Method/Year}} & RGB & IMU & Depth & Wheel & GNSS & LiDAR & Open-source &  Degradation-aware  & Mapping \\
        \hline
        
        VINS-RGBD\cite{shan2019rgbd}, 2019 & \Checkmark & \Checkmark & \Checkmark &  &  &  & \Checkmark &  & SPC$^1$/Non-color\\
        
        DRE-SLAM\cite{yang2019dre}, 2019 & \Checkmark &  & \Checkmark & \Checkmark &  &  & \Checkmark &  & Mesh/Non-color\\

        GR-Fusion\cite{wang2021gr}, 2021 & \Checkmark & \Checkmark &  &\Checkmark  &\Checkmark  & \Checkmark &  &  & SPC/Non-color\\

        LVI-SAM\cite{shan2021lvi}, 2021 & \Checkmark & \Checkmark &  &  &  & \Checkmark &  \Checkmark &  & SPC/Non-color\\

        VIW-Fusion\cite{Tingda2022VIW}, 2022 & \Checkmark & \Checkmark &  & \Checkmark &  &  & \Checkmark &  & SPC/Non-color\\
        
        R3LIVE\cite{lin2021r3live}, 2022 & \Checkmark & \Checkmark &  &  &  & \Checkmark &  \Checkmark &  & \fs \bf{Mesh/Color}\\

        DAMS-LIO\cite{han2023dams}, 2023 &  & \Checkmark &  & \Checkmark &  & \Checkmark &   & \Checkmark & {SPC/Non-color}\\

        M2C-GVIO\cite{hua2023m2c}, 2023 & \Checkmark & \Checkmark &  &  & \Checkmark &  &   &  & Sparse/Non-color\\

        Ground-Fusion\cite{yin2024ground}, 2024 & \Checkmark & \Checkmark & \Checkmark & \Checkmark & \Checkmark &  & \Checkmark & \Checkmark & \nd \sl{DPC$^2$/Color}\\

        LIGO\cite{he2025ligo}, 2025 &  & \Checkmark &  &  & \Checkmark & \Checkmark & \Checkmark & & SPC/Non-color\\

        \hline
        \textbf{Ground-Fusion++(Ours)}, 2025 & \Checkmark & \Checkmark & \Checkmark & \Checkmark & \Checkmark & \Checkmark & \Checkmark & \Checkmark & \fs \bf{Mesh/Color}\\

        \hline
        \multicolumn{10}{l}{\footnotesize{$^1$ SPC stands for Sparse Point Cloud,  $^2$ DPC stands for Dense Point Cloud.}}
    \end{tabular}
    \end{adjustbox}
    \vspace{-5mm}
\end{table*}

\subsection{Multi-sensor Fusion SLAM Algorithm}

Over the past decade, visual SLAM\cite{teed2021droid,campos2021orb} and visual-inertial odometry (VIO)\cite{von2022dm,qin2018vins} have achieved centimeter-level localization accuracy in small-scale indoor environments\cite{schubert2018tum,burri2016euroc}, while LiDAR-based SLAM and LiDAR-Inertial-Odometry (LIO)\cite{xu2022fast,shan2020lio,dellenbach2022ct} have demonstrated greater stability in large-scale outdoor scenarios. Recent advances have focused on multi-sensor fusion, particularly LiDAR-visual fusion\cite{lin2021r3live,zheng2022fast,zheng2024fast}, to enhance localization accuracy and mapping quality. Beyond LiDAR-vision fusion, integrating wheel odometry\cite{Tingda2022VIW,wu2017vins} into VIO system further improves robustness in featureless or low-light environments, and GNSS integration\cite{cao2022gvins,he2025ligo,hua2023m2c,jiang2024innovation} enables global drift-free pose estimation. We provide curated overviews of leading LiDAR-visual SLAM and wheel-integrated SLAM at \footnote{\href{https://github.com/sjtuyinjie/awesome-LiDAR-Visual-SLAM}{https://github.com/sjtuyinjie/awesome-LiDAR-Visual-SLAM}} and \footnote{\href{https://github.com/sjtuyinjie/awesome-Wheel-SLAM}{https://github.com/sjtuyinjie/awesome-Wheel-SLAM}}, respectively, which we do not elaborate on here.
Recent studies have revealed that these cutting-edge SLAM systems struggle with significant drift or even complete tracking failures under specific sensor degradation scenarios\cite{yin2023ground,yin2024ground}. In GNSS-denied areas, LiDAR degeneracy, or severe motion blur, conventional SLAM approaches often fail to initialize or lose tracking, severely limiting their reliability in real-world applications.

To mitigate these challenges, recent works\cite{han2023dams,yin2024ground,yin2023sky} have explored degeneration detection strategies aimed at improving system robustness. For example, Ground-Fusion\cite{yin2024ground} tightly integrates RGB-D, IMU, wheel odometry and GNSS, and employs an adaptive sensor selection strategy to enhance performance in corner cases. However, Ground-Fusion still faces challenges in ensuring long-term localization and mapping stability in large-scale outdoor environments, which is critical for downstream applications such as navigation and exploration\cite{yin2024disentangled}.

Building on these insights, we introduce Ground-Fusion++, a degradation-aware multi-sensor fusion SLAM framework that extends Ground-Fusion by incorporating LiDAR alongside RGB-D, IMU, wheel odometry, and GNSS. By leveraging an adaptive sensor selection strategy, Ground-Fusion++ enables resilient localization on long-term outdoor trajectories. Furthermore, Ground-Fusion++ achieves real-time dense colorized mapping by seamlessly incorporating advanced rendering techniques \cite{lin2023immesh}, effectively addressing the limitations of its predecessor. A detailed comparative analysis of these systems is provided in Table~\ref{slam_comparison}.

\section{M3DGR Benchmark Dataset Construction}
\vspace{-2mm}
\subsection{Acquisition Platform} 
As shown in Figure~\ref{robot_car}, we have constructed a ground robot equipped with multiple sensors to capture comprehensive datasets, including RGBD images, 3D point clouds, wheel odometry, and raw GNSS signals. Data acquisition and recording are managed by a high-performance Intel NUC with a high-speed NMVe SSD. 

\begin{figure*}[htbp]
    \centering
    \includegraphics[width=1.58\columnwidth]{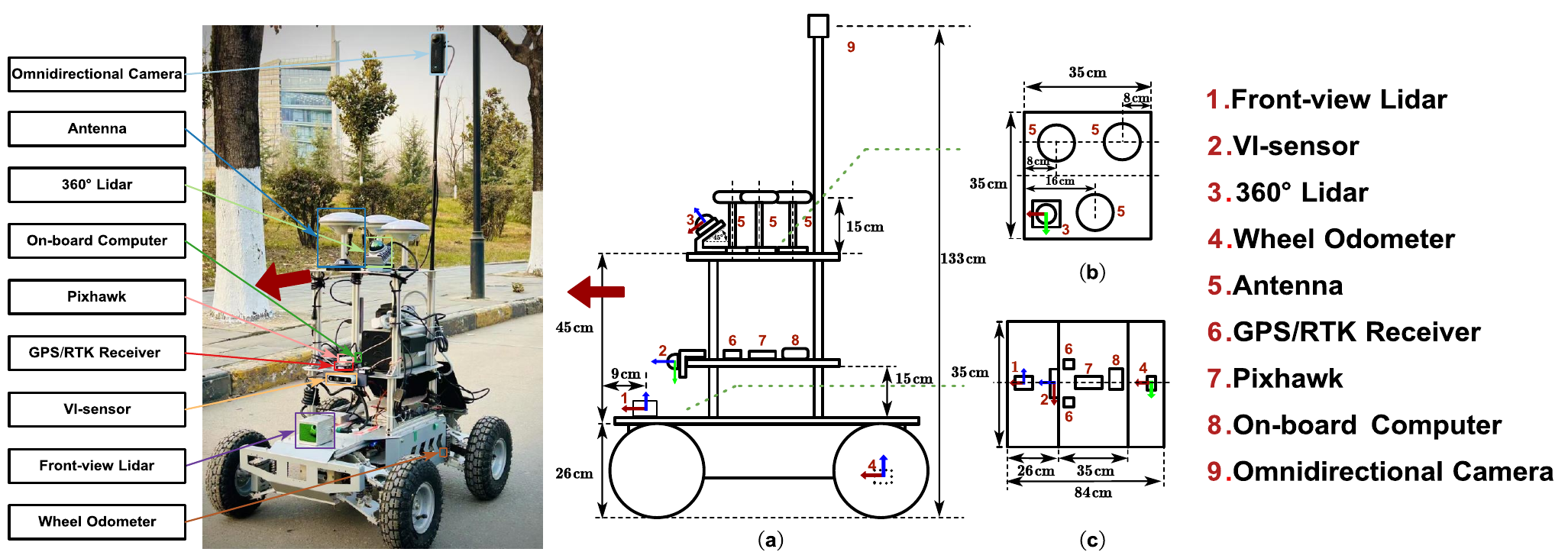}
    \caption{Physical drawings and schematics of the ground robot. (a) Side view of the robot. (b) Sensor arrangement on the top layer. (c) Sensor arrangement on the middle and bottom layers. All dimensions are provided in centimeters.} 
    \label{robot_car}
    \vspace{-4mm}
\end{figure*}

\subsection{Sensor Setup} 
The physical design and sensor layouts of the ground robot are illustrated in Figure \ref{robot_car}. The platform is equipped with a differential drive system, featuring two motorized wheels that provide wheel odometry data. For visual data collection, an RGBD-IMU sensor captures forward-facing images along with inertial measurements. Furthermore, two non-repetitive scanning Livox LiDARs are mounted to generate high-resolution 3D point clouds of the surrounding environment. To obtain raw GNSS signals, a GNSS receiver is equipped on the top of the robot. To ensure accurate ground-truth trajectories, high-precision trajectory data is obtained using a motion capture (Mocap) system for indoor experiments and
a Real-Time Kinematic (RTK) receiver in outdoor environments as shown in Figure \ref{dataset_explain}(b). A detailed overview of the specifications of all sensors and tracking devices is provided in Table \ref{m3dgr_sensor}.

 \begin{table}[ht]
        \small
        \caption{Sensor and Tracking Device Specifications}
        \renewcommand{\arraystretch}{1.3} 
        \label{m3dgr_sensor}
        \centering
        \begin{adjustbox}{width=1\columnwidth}
            \begin{tabular}{cccc}
                \hline
                \makecell{Device} & \makecell{Type} & \makecell{Spec.} & \makecell{Freq.(Hz)} \\               
                \hline
                \multirow{3}*{RGBD-IMU sensor} & \multirow{3}*{Realsense D435i} &  RGB: 640*480, 69°(H) × 42°(H) & 30 \\
                & & Depth: 640*480, 87°(H) × 58°(V) & 30 \\
                & & IMU: 6-axis & 200 \\
                Omnidirectional Camera & Insta360 X4 & RGB: 2880x1440, 360° & 15 \\
                
                \multirow{4}*{LiDAR} & \multirow{2}*{Livox Avia} & Non-repetitive, 450m, 70.4°(H) × 77.2°(V) & 15 \\
                    & & IMU(BMI088): 6-axis & 200\\ 
                    & \multirow{2}*{Livox MID360} & Non-repetitive, 40m,  360°(H) × 59°(V) & 15 \\
                    & & IMU(ICM40609): 6-axis & 200\\ 
                Wheel Odometer & WHEELTEC & 2D & 20 \\
                GNSS Receiver  & CUAV C-RTK 9Ps & BDS/GPS/GLONASS/Galileo & 10 \\

                \hline
                RTK Receiver & CUAV C-RTK 2HP & localization accuracy 0.8cm(H)/1.5cm(V) & 15 \\
                Mocap System & Optitrack & localization accuracy 1mm & 360 \\
                \hline
            \end{tabular}
        \end{adjustbox}
        \vspace{-3.5mm}
    \end{table}

\subsection{Synchronization}
We use the rosbag tool in the Robot Operating System (ROS) to record data from all sensors, ensuring a unified timestamp mechanism for synchronization. Some sensors feature built-in hardware synchronization capabilities. For example, the RealSense D435i, which integrates an RGBD camera and a 6-axis IMU, achieves internal synchronization by simultaneously triggering data acquisition. Similarly, the two Livox LiDARs, each equipped with a 6-axis IMU, employ hardware triggering to capture data at the same instance, ensuring internal synchronization.

    \begin{figure}[!htp]
        \small
        \begin{center}
            \begin{tabular}{cc}
            \includegraphics[scale=0.17]{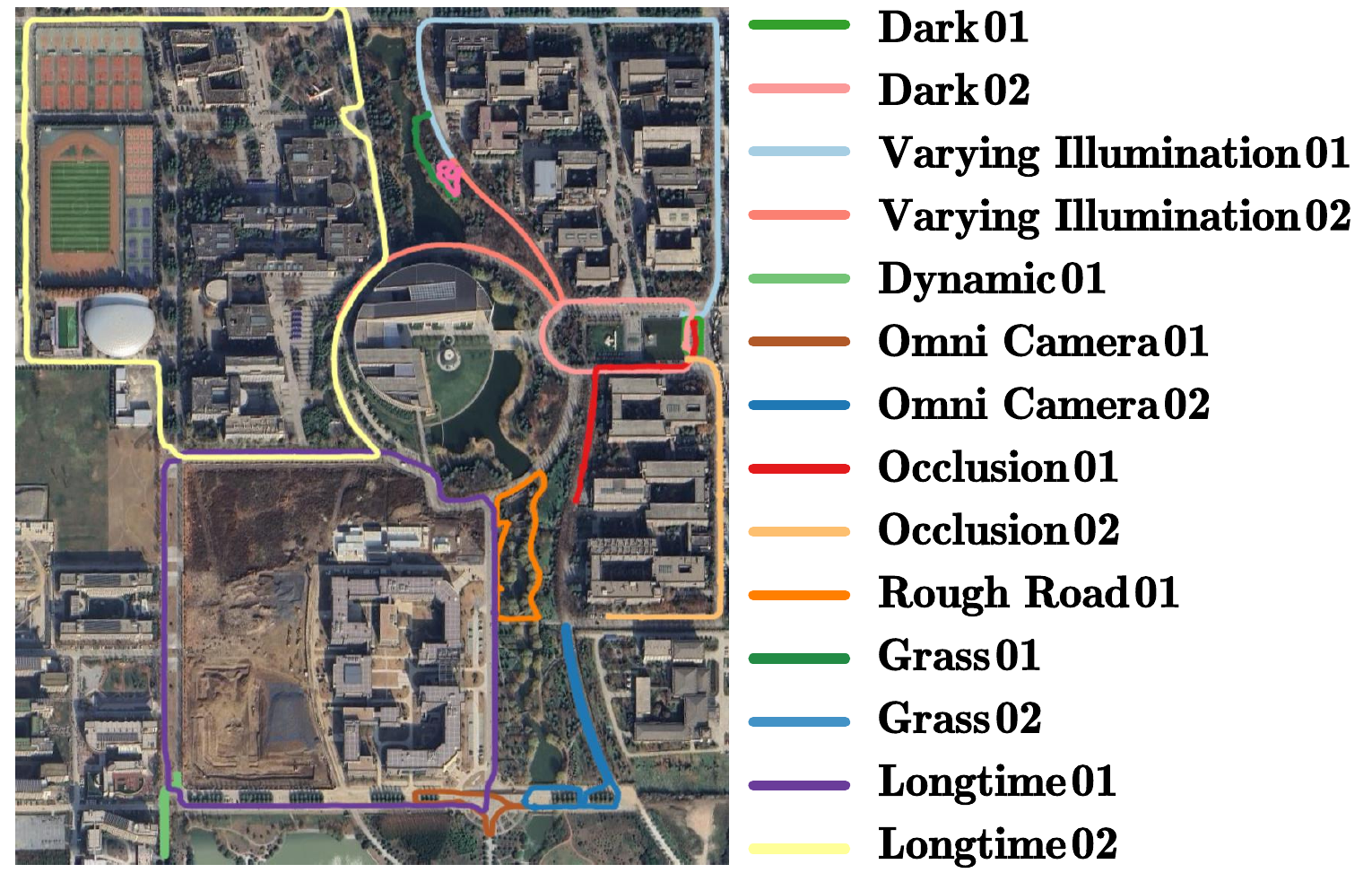} &
            \includegraphics[scale=0.17]{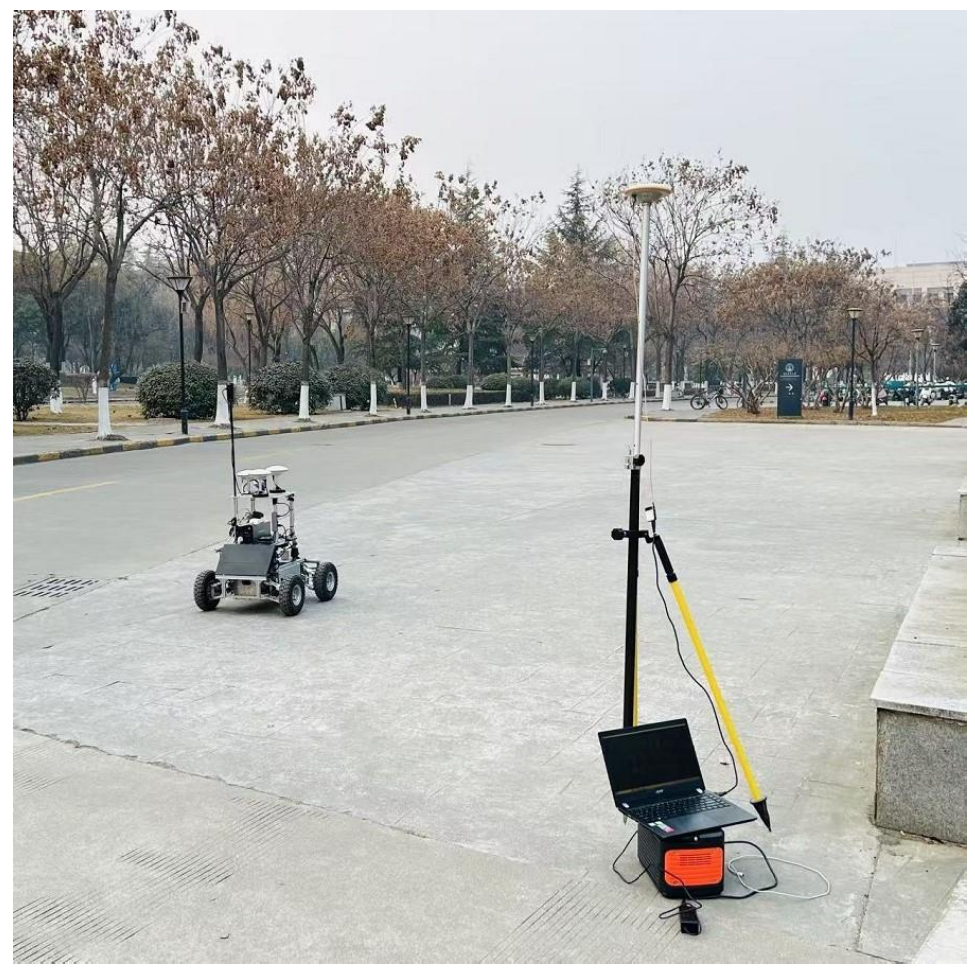}  \\
            (a) & (b)  \\
            \end{tabular}
        \end{center}
        \vspace{-4mm}
        \caption{(a) We visualize the trajectories of outdoor sequences in the map with different colors. (b) The ground truth for outdoor sequences was obtained using a geodetic-grade GNSS receiver in RTK mode, ensuring centimeter-level positioning accuracy.}
        \label{dataset_explain}
        \vspace{-5mm}
    \end{figure}

    \begin{figure}
        \small
        \begin{center}
            \begin{tabular}{cccc}
            \includegraphics[scale=0.099]{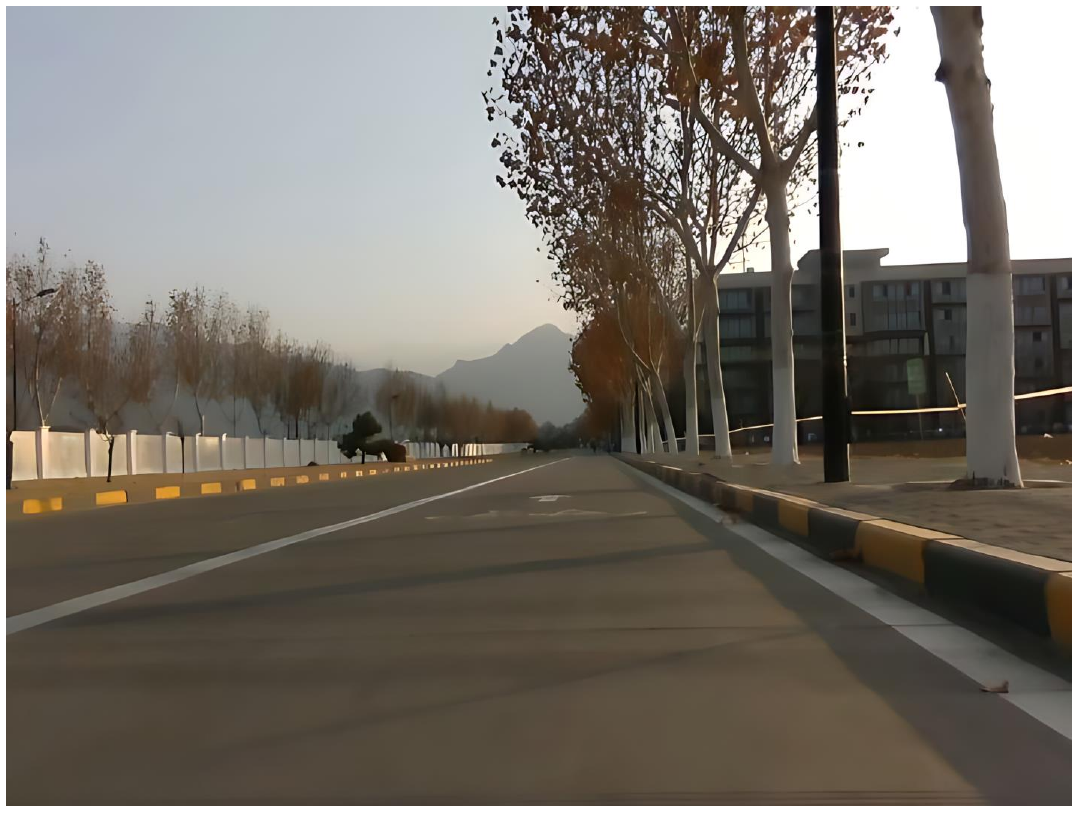} &
            \includegraphics[scale=0.099]{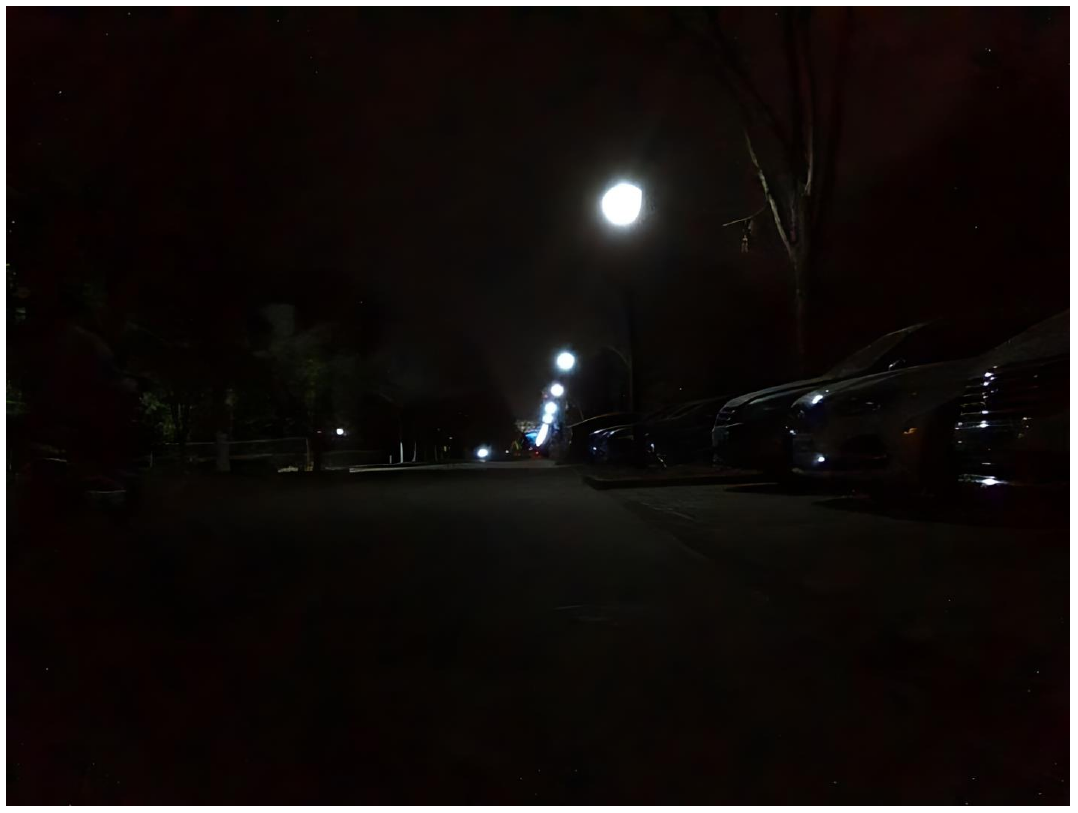}  &
            \includegraphics[scale=0.099]{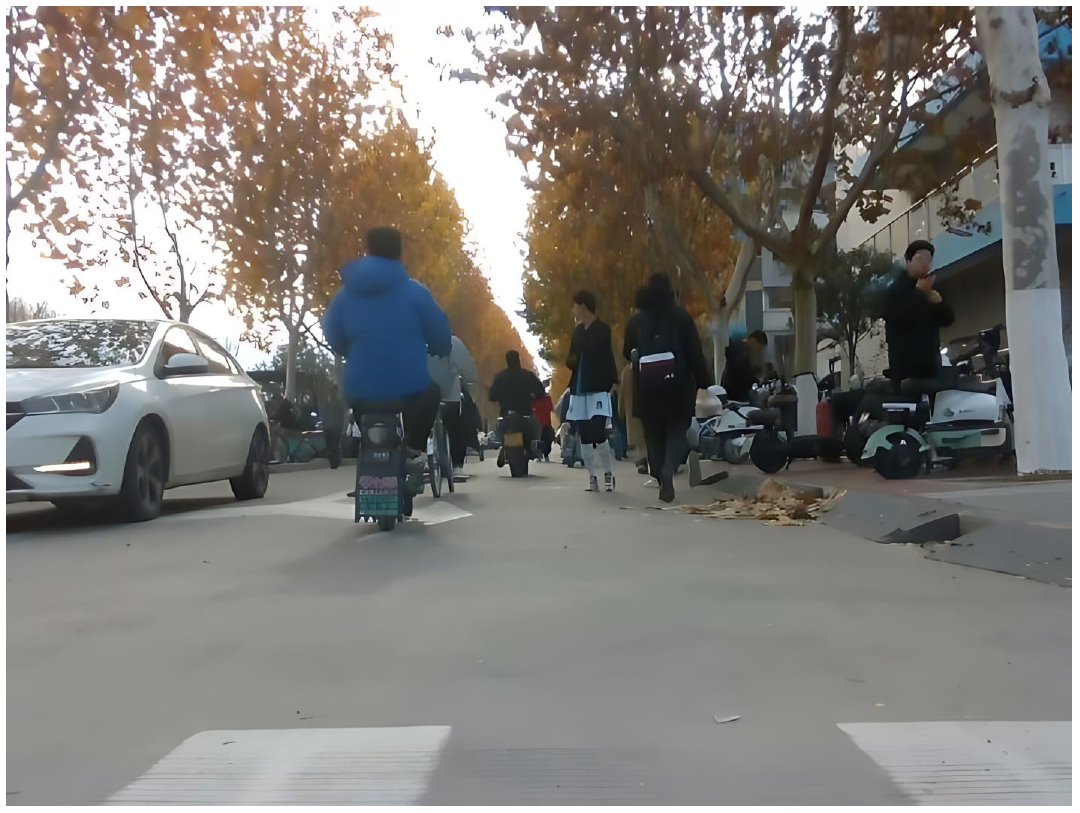} &
            \includegraphics[scale=0.099]{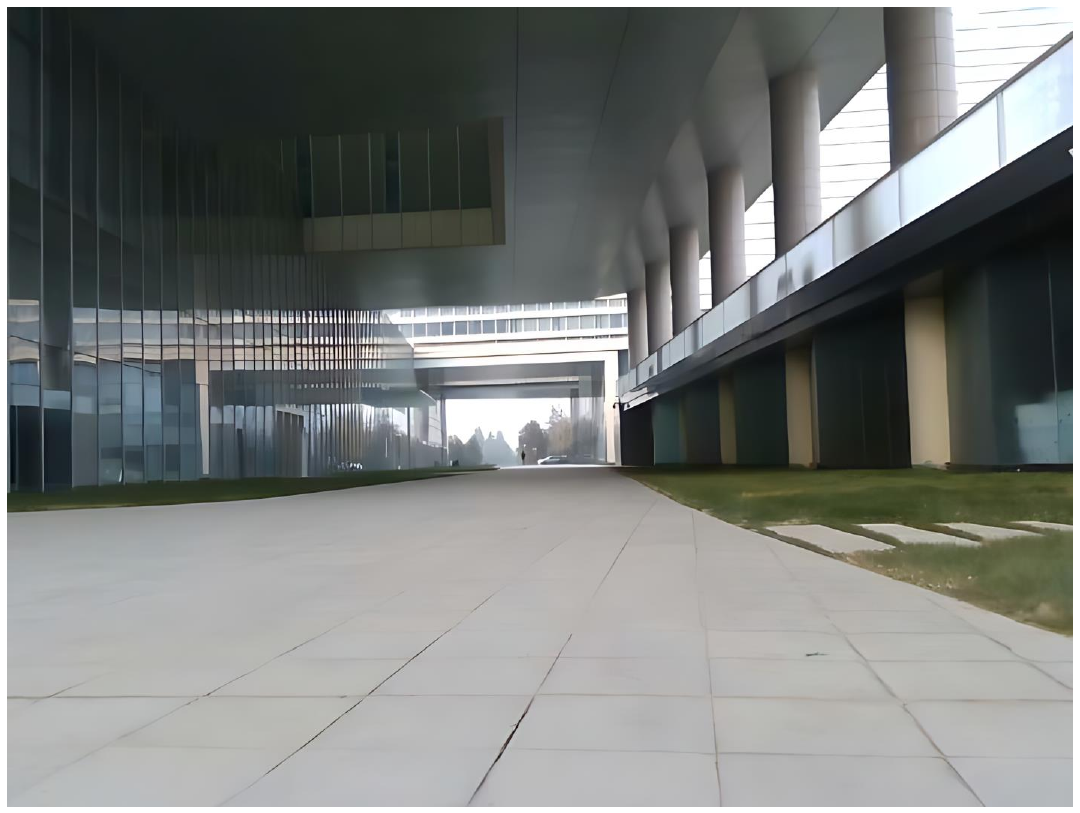} \\
            (a) & (b) & (c) & (d) \\

            \includegraphics[scale=0.099]{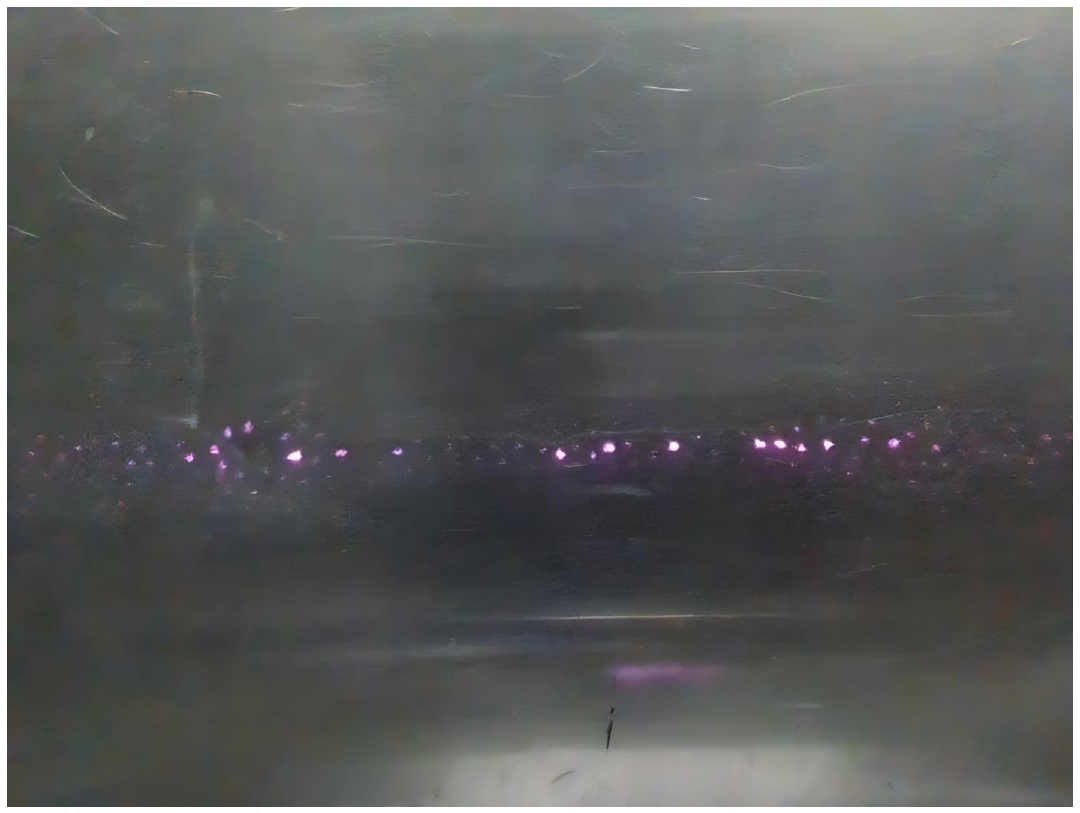}     &
            \includegraphics[scale=0.099]{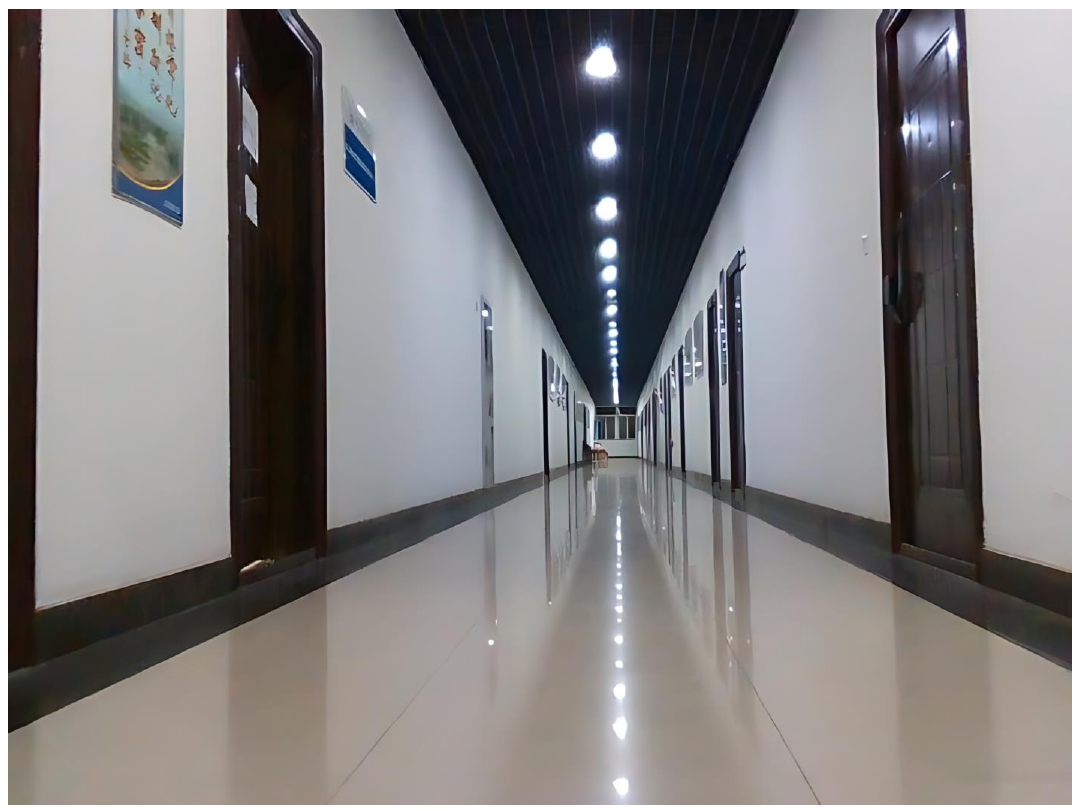} & 
            \includegraphics[scale=0.099]{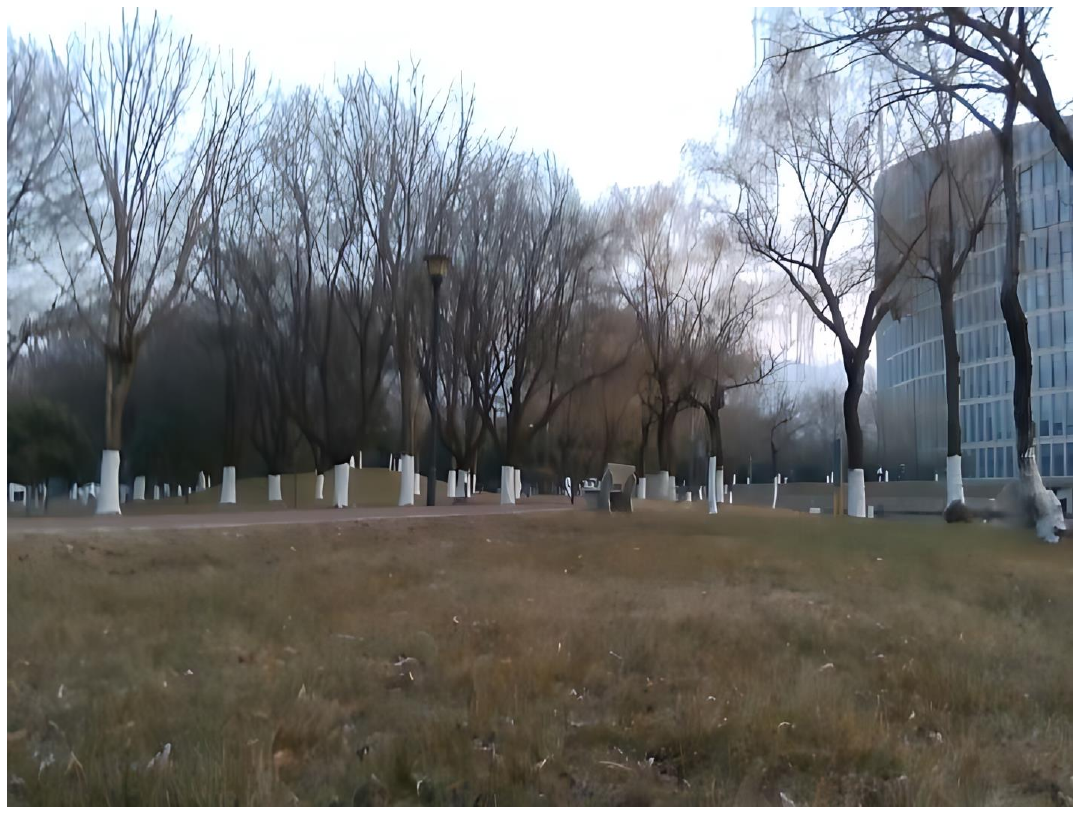}    & 
            \includegraphics[scale=0.099]{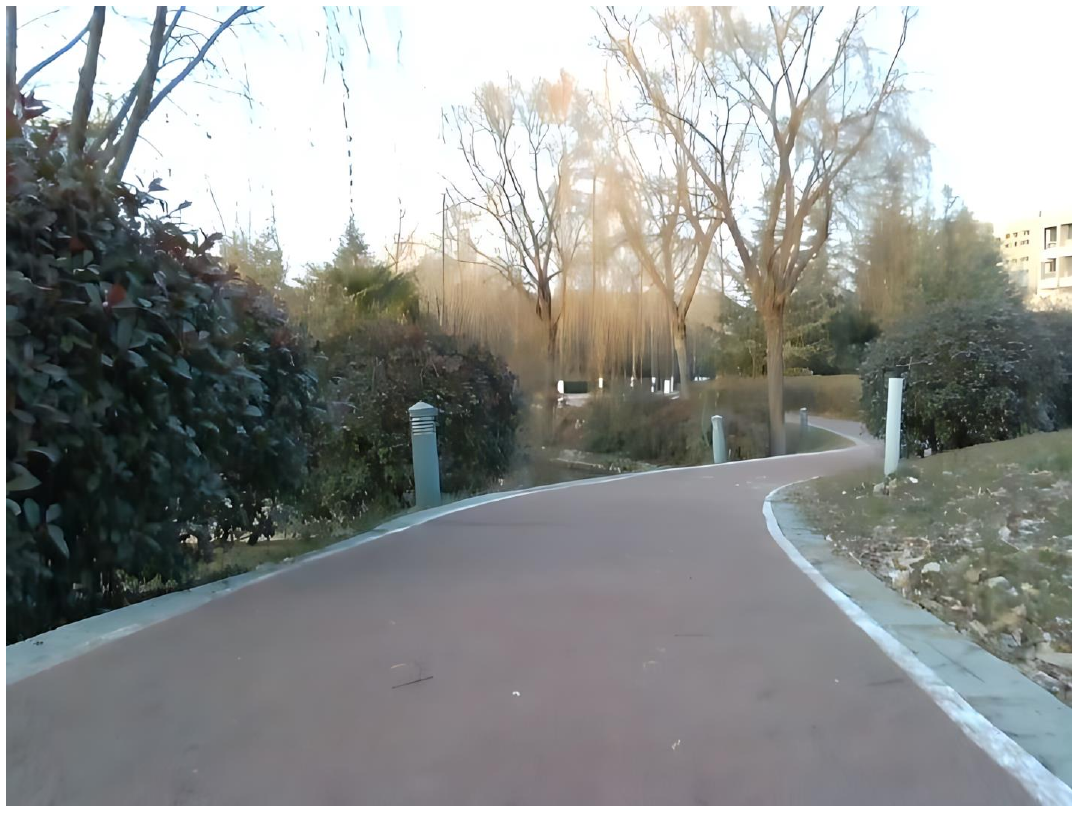} \\
            (e) & (f) & (g) & (h) \\
            \end{tabular}
        \end{center}
        \vspace{-4mm}
        \caption{Diverse scenarios included in our datasets: (a) Normal lighting conditions. (b) Low-light environment. (c) A park with significant pedestrian dynamics. (d) An area without GNSS signal coverage. (e) An elevator. (f) A narrow corridor. (g) A grassy terrain. (h) A road with abrupt elevation changes.}
        \vspace{-5mm}
        \label{corner_case}
    \end{figure}

\begin{table*}[ht]
        \small
        \caption{An overview of scenarios in M3DGR dataset.}
        \centering
        \renewcommand{\arraystretch}{1.3}
        \label{overview_tab}
        \begin{adjustbox}{width=1.9\columnwidth}
        \begin{tabular}{*{16}c}
            \hline
            \multirow{2}{*}{\makecell{\textbf{Scenario}}} 
            & \multicolumn{5}{c}{\makecell{Visual Challenge}} 
            & \multicolumn{2}{c}{\makecell{LiDAR Degeneracy}}
            &
            & \multicolumn{4}{c}{\makecell{Wheel Slippage}}  
            & \multirow{2}{*}{\makecell{GNSS Denial}} & 
            \multirow{2}{*}{\makecell{Standard} } & \multirow{2}{*}{\makecell{TOTAL}} \\ 
                \cline{2-5} \cline{7-8} \cline{10-13} 
                & Dark & VI$^1$ & Dynamic & Occlusion &  & Corridor & Elevator & & WF$^2$ & ST$^3$ & Grass & RR$^4$ & \\   
            \hline
            \makecell{\textbf{Number}} & 5 & 4 & 3 & 4 &  & 2 & 1  & & 2 & 2 & 2 & 1 &  2 & 4 & 32 \\
            \makecell{\textbf{Dist/m}} & 1653.31 & 1055.58 & 355.97 & 1091.24 &  & 545.64 & 470.64 & & 101.55 & 170.88 & 318.91 & 457.35 & 1162.39 & 4485.49 & 11868.95 \\
            \makecell{\textbf{Duration/s}} & 2274 & 1458 & 609 & 1224 &  & 696 & 699 & & 171 & 238 & 459 & 533 & 1359 & 5101 & 14821 \\
            \makecell{\textbf{Size/GB}} & 27.0 & 20.0 & 7.1 & 12.3 &  & 11.9 & 11.2 & & 3.3 & 2.9 & 9.7 & 10.4 & 23.2 & 86.0 & 225.0 \\
            \makecell{\textbf{GroundTruth}} & RTK/Mocap & RTK/Mocap & RTK/Mocap & RTK/Mocap & & ArUco & ArUco & & Mocap & Mocap & RTK & RTK & ArUco & RTK & ---- \\
            \hline
            \multicolumn{15}{l}{\footnotesize{$^1$ stands for varying illumination, $^2$ stands for wheel float, $^3$ stands for sharp turn, $^4$ stands for rough road.}}
        \end{tabular}
        \end{adjustbox}
        \vskip 1mm
        \vspace{-6.5mm}
    \end{table*}

\subsection{Data collection}
We deploy our ground robot in both standard and challenging environments to collect a diverse dataset. An overview of the recorded sequences is presented in Table \ref{overview_tab}. The corner cases in our dataset are classified into four categories: visual challenges, LiDAR degeneracy, wheel slippage, and GNSS-denied conditions, covering both indoor and outdoor scenarios. In addition to these challenging cases designed to test system robustness, our dataset also includes representative standard scenarios for evaluation. 
To provide an overview of outdoor sequences, Figure \ref{dataset_explain}(a) illustrates the recorded trajectories on a map. 

\par \textbf{Visual challenge} The dataset includes four types of visual challenges: low-light conditions, varying illumination, dynamic scenes, and occlusion. These scenarios are designed to test the resilience of vision-based perception systems in both indoor and outdoor environments.
For indoor sequences, low-light conditions are simulated by turning off the room lights and using only a mobile phone flashlight for illumination. Varying illumination is introduced by alternating the room lights at regular intervals. To create dynamic scenes, a person moves continuously in front of the ground robot, disrupting its field of view. Additionally, occlusion scenarios are generated by deliberately blocking the camera.
For outdoor sequences, low-light and varying illumination challenges are captured during nighttime operations. Low-light scenarios are recorded in dark environments, as shown in Figure \ref{corner_case}(b), while varying illumination sequences are obtained by navigating the robot through areas with different lighting conditions. Dynamic environments are simulated by recording sequences with multiple pedestrians, cyclists, and moving vehicles, as illustrated in Figure \ref{corner_case}(c). To induce occlusion, the camera is partially obstructed at various points during data collection.

\par \textbf{LiDAR degeneracy} The dataset includes two LiDAR degeneracy scenarios: a long corridor and an elevator transition.
In the first scenario, the ground robot navigates through a long corridor, where the lack of geometric features poses challenges for LiDAR-based localization, as illustrated in Figure \ref{corner_case}(f). The second scenario involves the robot moving from a long corridor into an elevator, introducing a sudden shift from a structured environment to a feature-sparse vertical space, as shown in Figure \ref{corner_case}(e).
To assess SLAM algorithm performance under these conditions, the robot is operated along a loop, returning to its starting position. The relative transformation between the initial and final frames, computed using ArUco marker detection, serves as the quantitative metric for evaluating trajectory accuracy.

\par \textbf{Wheel slippage} The dataset captures four distinct wheel slippage scenarios: wheel float, sharp turns, grass-covered surfaces, and rough roads.
Wheel float is induced by lifting the robot’s driving wheels, causing intentional wheel slippage. Sharp turns are executed by commanding the robot to make rapid directional changes at high speeds, highlighting dynamic slippage behavior. Grass-covered surfaces introduce varying traction conditions as the robot navigates through uneven terrain, as shown in Figure \ref{corner_case}(g). Rough roads, characterized by significant elevation changes, further challenge the robot’s mobility, as illustrated in Figure \ref{corner_case}(h).

\par \textbf{GNSS denial} The dataset includes a GNSS-denied scenario in which the ground robot begins in a GNSS-available area, traverses a GNSS-denied region (illustrated in Figure \ref{corner_case}(d)), and then returns to its starting point. To evaluate localization performance, the relative transformation between the start and end positions is measured using ArUco marker detection, as previously described

\vspace{-2mm}


\section{System Design of Ground-Fusion++}

To provide a resilient baseline system for the M3DGR benchmark, we develop Ground-Fusion++, a modular and degradation-aware multi-sensor fusion framework. Inspired by\cite{shan2021lvi,han2023dams}, Ground-Fusion++ avoids a fully tightly coupled multi-sensor fusion approach and instead employs a hybrid framework consisting of an enhanced VIO subsystem and a LIO subsystem. We adopt Ground-Fusion\cite{yin2024ground} as the powerful visual odometry module and utilize a continuous-time version of Fast-LIO2\cite{xu2022fast}, which leverages advancements from \cite{chengwei2023CT}, as the LIO subsystem. 

Our system follows a LiDAR-priority selection strategy, where LIO is the primary source of pose estimation unless LiDAR degradation is detected. If LiDAR degeneration occurs, the system adaptively switches to the enhanced VIO subsystem (Ground-Fusion), provided it can be successfully initialized. The VIO subsystem retains degradation detection and adaptive sensor selection strategies from Ground-Fusion\cite{yin2024ground}, leveraging RGBD, GNSS, IMU and wheel odometry to ensure robustness. 

To determine whether LiDAR degradation occurs, we evaluate the following condition:
\begin{equation}
\vspace{-1mm}
 \scriptsize
\text{Deg}(t) = 
\begin{cases} 
1, & \text{if } N_t^{\text{feat}} < \tau_N \ \text{OR} \ \epsilon_t^{\text{align}} > \tau_\epsilon \\
0, & \text{otherwise}
\end{cases}
\label{eq:degradation}
\end{equation}
where $N_t^{\text{feat}}$ denotes the LiDAR feature count at time $t$, $\epsilon_t^{\text{align}} = \frac{1}{N}\sum_{i=1}^N \|\bm{p}_i - \bm{q}_i\|^2$ represents the ICP alignment residual with $\bm{p}_i$ and $\bm{q}_i$ being corresponding point pairs between consecutive scans, $N$ is the total number of matched points, and $\tau_N$, $\tau_\epsilon$ are empirical thresholds for feature count and alignment error respectively.

To ensure consistent frame alignment between subsystems, we formulate the coordinate transformation optimization as follows:
\vspace{-2mm}

\begin{equation}
\vspace{-2mm}
 \scriptsize
\bm{T}_{\text{align}} = \mathop{\arg\min}\limits_{\bm{T} \in SE(3)} \sum_{k=1}^K \rho\left(\left\| \log\left(\bm{T}_{\text{LIO}}^k (\bm{T}_{\text{VIO}}^k)^{-1}\right) \right\|_{\bm{\Sigma}_k^{-1}}^2 \right)
\label{eq:alignment}
\end{equation}
where
$\bm{T}_{\text{align}} \in SE(3)$ represents the optimal coordinate transformation from LIO frame to VIO frame,
$\bm{T}_{\text{LIO}}^k, \bm{T}_{\text{VIO}}^k \in SE(3)$ denote pose estimates from LIO and VIO subsystems at time $k$,
$\bm{\Sigma}_k \in \mathbb{R}^{6\times6}$ is the uncertainty covariance matrix.
$\rho(\cdot)$ denotes the Cauchy robust kernel.
$\|\bm{v}\|_{\bm{\Sigma}^{-1}}$ is the Mahalanobis norm.

For smooth trajectory transitions, we use an exponential map-based correction:
\begin{equation}
\scriptsize
\vspace{-2mm}
\bm{T}_t^{\text{fused}} = \bm{T}_t^{\text{active}} \exp\left(-\beta \left[\log\left( (\bm{T}_t^{\text{active}})^{-1} \bm{T}_{\text{align}} \bm{T}_t^{\text{backup}} \right)\right]^\wedge\right)
\label{eq:smoothing}
\end{equation}
where
$\bm{T}_t^{\text{active}}/\bm{T}_t^{\text{backup}} \in SE(3)$ are poses from active/backup subsystems.
 $\beta \in [0,1]$ is the smoothing factor.

Finally, the RGB images, LiDAR point clouds, and odometry outputs from the active subsystem are fed into the rendering module \cite{lin2023immesh} to generate real-time, dense, colorized maps. Although Ground-Fusion++ primarily builds upon existing frameworks, \textbf{its modular design and comprehensive sensor integration ensure strong scalability and flexibility}. In particular, the carefully designed degradation detection mechanism and trajectory smoothing strategy effectively mitigate the impact of sensor failures and subsystem switching, ensuring robust and seamless localization performance. This architecture allows the LIO subsystem, VIO subsystem, and rendering module to be seamlessly upgraded with the latest advancements, establishing an adaptable baseline for future developments.

    \begin{table*}[h]
        \small
        \centering
        \caption{Sample sequences for evaluation.}
        \renewcommand{\arraystretch}{1.3} 
        \label{sample_tab}
        \begin{adjustbox}{width=1.7\columnwidth}

        \begin{tabular}{*{11}c}
            \hline
            \makecell{\textbf{Sequence}} & Dynamic01 & Varying-illu01 & Occlusion01 & Dark01 & Corridor01 & Elevator01 & Wheel-float01 & Sha-turn01 & Grass01  & GNSS-denial01  \\
            \hline
            \makecell{\textbf{Duration/s}} & 175 & 154 & 142 & 206 & 403 & 699 & 123 & 138 & 287 & 609 \\
            \makecell{\textbf{Dist/m}} & 72.22 & 73.69 & 65.00 & 112.94 & 332.32 & 470.64 & 45.01 & 98.31 & 189.45 & 524.37 \\
            \makecell{\textbf{Speed/s}} & 0.41 & 0.47 & 0.45 & 0.55 & 0.82 & 0.67 & 0.36 & 0.71 & 0.66 & 0.86 \\
            \makecell{\textbf{Discription}\\\textbf{of features}} & \makecell{indoor, dynamic\\pedestrians} & \makecell{indoor, varying\\light conditions} & \makecell{indoor, complete\\visual occlusion} & \makecell{outdoor, low\\light conditions} & \makecell{indoor,\\long corridor} & \makecell{indoor, long\\corridor with Elevator} & \makecell{indoor,\\tire slip} & \makecell{outdoor,\\quickly turn} & \makecell{outdoor, grass\\covered surface} & \makecell{outdoor, GNSS\\denied zones}   \\
            \hline
        \end{tabular}
        \end{adjustbox}
        \vspace{-4mm}
    \end{table*}

\section{Experiments and Evaluations}

\begin{table*}[ht]
    \caption{ATE RMSE(m) of SLAM systems on M3DGR sequences.}
    \label{rmse_tab}
    \renewcommand{\arraystretch}{1.3} 
    \begin{adjustbox}{width=2\columnwidth}
    \centering
    \vspace{-1.5mm}
    \begin{tabular}{*{15}c}
        \hline
        \multirow{2}{*}{\makecell{{Method/Scenario}}} 
          & \multicolumn{4}{c}{\makecell{Visual Challenge}} 
          &
          & \multicolumn{2}{c}{\makecell{LiDAR Degeneration}} 
          &
          & \multicolumn{3}{c}{\makecell{Wheel Slippage}} 
          &
          & \multicolumn{1}{c}{\makecell{GNSS Denial}} \\ 
        \cline{2-5}  \cline{7-8} \cline{10-12} \cline{14-14}
        & Dynamic01 & Varying-illu01 & Dark01 & Occlusion01 & & Corridor01 & Elevator01 & & Wheel-float01 & Sha-turn01 & Grass01 & & GNSS-denial01\\
        \hline
       
       Wheel Odom & \makecell[c]{2.32} & \makecell[c]{2.36} & \makecell[c]{5.52} & \makecell[c]{2.04} & & \makecell[c]{72.61} & \makecell[c]{66.94} & & \makecell[c]{2.20} & \makecell[c]{7.44} & \makecell[c]{25.34} & & \blackx\\ 
       
       GNSS SPP & \blackx & \blackx & \makecell[c]{7.69} & \blackx & & \blackx & \blackx & & \blackx & \blackx & \makecell[c]{0.48} & & \makecell[c]{11.61}\\
        \hline   
       TartanVO\cite{wang2021tartanvo}, 2021  & \makecell[c]{2.37} & \makecell[c]{2.17} & \makecell[c]{12.37} & \blackx & & \blackx & \blackx & & \makecell[c]{1.93} & \makecell[c]{2.09} & \makecell[c]{4.68} & & \blackx \\
       
       ORB-SLAM2\cite{mur2017orb}, 2017 & \fs \bf{0.14} & \blackx & \blackx & \blackx & & \makecell[c]{6.41} & \rd{8.09} & & \makecell[c]{1.72} & \makecell[c]{1.54} & \blackx & & \blackx \\
       
       
       ORB-SLAM3\cite{campos2021orb}, 2021 & \blackx & \blackx & \blackx & \blackx & & \blackx & \blackx & & \blackx & \blackx & \blackx & & \blackx\\
       
       DM-VIO\cite{von2022dm}, 2022 & \makecell[c]{2.25} & \makecell[c]{2.27} & \rd{4.08} & \blackx & & \makecell[c]{12.20} & \fs \bf{2.54} & & \blackx & \makecell[c]{8.90} & \blackx & & \blackx\\
       
       VINS-Mono\cite{qin2018vins}, 2018 & \makecell[c]{0.43} & \makecell[c]{2.70} & \makecell[c]{7.91} & \blackx & & \makecell[c]{9.82} & \makecell[c]{62.80} & & {0.46} & \nd \sl{0.36} & \nd \sl{2.17} & & \makecell[c]{30.36} \\
       
       VINS-RGBD\cite{shan2019rgbd}, 2019 & \rd{0.20} & \makecell[c]{1.86} & \blackx & \blackx & & \rd{5.62} & \blackx & & \nd \sl{0.28} & \fs \bf{0.35} & \makecell[c]{7.52} & & \rd{26.31} \\
       
       GVINS\cite{cao2022gvins}, 2022  & \makecell[c]{0.26} & \rd{1.25} & \blackx & \blackx & & \makecell[c]{9.42} & \nd \sl{2.89} & & \fs \bf{0.27}  & \rd{0.40} & \blackx & & \blackx \\
       
       VIW-Fusion\cite{Tingda2022VIW}, 2022 & \makecell[c]{0.62} & \nd \sl{1.02} & \fs \bf{0.77} & \blackx & & \nd \sl{5.58} & \makecell[c]{16.68} & & \makecell[c]{0.77} & \makecell[c]{2.44} & \rd{2.91} & & \makecell[c]{99.06}\\
       
       VINS-GPS-Wheel\cite{Wallong2021}, 2021  & \makecell[c]{1.18} & \makecell[c]{1.32} & \makecell[c]{15.55} & \blackx & & \fs \bf{5.55} & \makecell[c]{43.48} & & \makecell[c]{0.86} & \makecell[c]{2.00} & \makecell[c]{18.47} & & \nd \sl{16.89}\\
       
       Ground-Fusion\cite{yin2024ground}, 2024  & \nd \sl{0.19} & \bf \fs{0.59} & \nd \sl{1.10} & \fs \bf{1.21} & & \makecell[c]{26.25} & \makecell[c]{29.93} & & \rd{0.29} & \makecell[c]{1.16} & \fs \bf{1.33} & & \fs \bf{13.19}\\ 
       
       \hline
       A-LOAM\cite{qin2018aloam}, 2018 & \makecell[c]{0.15} & \makecell[c]{0.16} & \makecell[c]{6.36} & \makecell[c]{0.19} & & \makecell[c]{66.66} & \makecell[c]{48.37} & & \makecell[c]{0.29} & \makecell[c]{0.26} & \makecell[c]{1.92} & & \makecell[c]{8.46}\\
       
       LOAM-Livox\cite{lin2020loam}, 2020 & \makecell[c]{2.88} & \makecell[c]{3.24} & \makecell[c]{2.55} & \makecell[c]{2.42} & & \makecell[c]{43.91} & \makecell[c]{87.52} & & \makecell[c]{1.47} & \makecell[c]{1.78} & \makecell[c]{4.36} & & \blackx\\
       
       CTLO\cite{chengwei2023CTLO}, 2023 & \fs \bf{0.10} & \rd{0.12} & \fs \bf{0.15} & \makecell[c]{0.14} & & \makecell[c]{3.29} & \makecell[c]{52.31} & & \makecell[c]{0.18} & \makecell[c]{0.15} & \makecell[c]{1.34} & & \makecell[c]{0.88}\\
       
       LeGO-LOAM\cite{shan2018lego}, 2018  & \makecell[c]{7.92} & \blackx & \makecell[c]{13.40} & \makecell[c]{6.28} & & {19.65} & \blackx & & \makecell[c]{5.89} & \makecell[c]{8.40} & \makecell[c]{32.66} & & \makecell[c]{95.91}\\
       
       LIO-mapping\cite{ye2019tightly}, 2019  & \makecell[c]{2.03} & \makecell[c]{1.95} & \blackx & \makecell[c]{2.46} & & \blackx & \blackx & & \makecell[c]{1.05} & \makecell[c]{1.46} & \makecell[c]{9.75} & & \blackx\\
       
       LIO-SAM\cite{shan2020lio}, 2020 & \makecell[c]{5.10} & \makecell[c]{2.24} & \blackx & \makecell[c]{1.31} & & {36.15} & \blackx & & \makecell[c]{0.73} & \makecell[c]{0.63} & {1.34} & & \makecell[c]{43.21}\\
       
       LINS\cite{qin2020lins}, 2020  & \makecell[c]{10.18} & \makecell[c]{3.40} & \makecell[c]{13.25} & \makecell[c]{5.07} & & \blackx & \blackx & & \makecell[c]{5.03} & \makecell[c]{5.04} & \makecell[c]{87.04} & & \blackx \\
       
       LiLi-OM\cite{li2021towards}, 2021  & \makecell[c]{1.49} & \makecell[c]{0.19} & \blackx & \makecell[c]{7.00} & & \blackx & \blackx & & \makecell[c]{0.35} & \makecell[c]{2.08} & \makecell[c]{2.22} & & \makecell[c]{15.96}\\
       
       LIO-Livox\cite{Livox2021LIO}, 2021 & \makecell[c]{0.18} & \makecell[c]{0.72} & \makecell[c]{0.30} & \makecell[c]{0.47} & & \blackx & \blackx & & \makecell[c]{0.35} & \makecell[c]{11.25} & \nd \sl{1.32} & & {6.63}\\

       Faster-LIO\cite{bai2022faster}, 2022 & \nd \sl {0.12} & {0.13} & \rd{0.17} & \fs \bf{0.11} & & \blackx & \blackx & & \makecell[c]{2.19} & \makecell[c]{2.84} & \makecell[c]{1.33} & & {5.60}\\
       
       IESKF-LIO\cite{chengwei2022iESKFlio}, 2022 & \makecell[c]{0.14} & \makecell[c]{0.14} & \fs \bf{0.15} & \nd \sl{0.13} & & \makecell[c]{14.38} & \blackx & & \makecell[c]{0.17} & \rd {0.16} & \nd \sl{1.32} & & \nd \sl{0.08}\\
       
       VoxelMap\cite{yuan2022efficient}, 2022 & \makecell[c]{0.89} & \makecell[c]{0.76} & \makecell[c]{4.93} & \makecell[c]{0.91} & & \fs \bf{1.08} & \nd \sl{19.22} & & \makecell[c]{0.99} & \makecell[c]{1.11} & \fs \bf{0.78} & & \fs \bf{0.07}\\
       
      Fast-LIO2\cite{xu2022fast}, 2022  & \rd{0.13} & \nd \sl{0.11} & {0.24} & \fs \bf{0.11} & & \blackx & \blackx & & \rd{0.16} & {0.18} & \nd \sl{1.32} & & \makecell[c]{13.46}\\

     Point-LIO\cite{he2023point}, 2023  & {0.14} & \makecell[c]{0.14} & \makecell[c]{0.29} & {0.15} & & {9.58} & \blackx & & {0.19} & \makecell[c]{0.20} & \nd \sl{1.32} & & \makecell[c]{18.39}\\
      
       LOG-LIO\cite{huang2023log}, 2023 & \rd{0.13} & \rd{0.12} & \makecell[c]{0.99} & \rd{0.14} & & \blackx & \blackx & & {0.18} & \fs \bf{0.07} & \rd{1.33} & & \makecell[c]{27.03}\\
       
       CT-LIO\cite{chengwei2023CT}, 2023 & \nd \sl{0.12} & \makecell[c]{0.12} & \makecell[c]{0.18} & \nd \sl{0.13} & & \rd{3.56} & \fs \bf{2.39} & & \fs \bf{0.13} & \nd \sl{0.10} & \makecell[c]{1.34} & & \makecell[c]{2.71}\\
       
       DLIO\cite{chen2023direct}, 2023 & \nd \sl{0.12} & \fs\bf{0.10} & \nd \sl{0.16} & {0.15} & & \makecell[c]{40.46} & {44.70} & & {0.18} & {0.17} & \makecell[c]{1.35} & & {5.80}\\
       
       HM-LIO\cite{chengwei2023hmlio}, 2023 & \nd \sl{0.12} & \makecell[c]{0.15} & \rd{0.17} & \rd{0.14} & & \nd \sl{3.23} & \blackx & & \rd{0.14} & \makecell[c]{0.20} & \rd{1.33} & & \rd{0.09}\\
       
       MM-LINS\cite{ma2024mm}, 2024  & \makecell[c]{2.84} & \makecell[c]{2.79} & \makecell[c]{0.27} & \makecell[c]{1.95} & & \blackx & \makecell[c]{74.31} & & \makecell[c]{2.25} & \makecell[c]{2.83} & \makecell[c]{1.67} & & \makecell[c]{7.73}\\
       
       LIGO\cite{he2025ligo}, 2025 & \fs \bf{0.10} &\rd{0.12} & \makecell[c]{0.25} & \makecell[c]{0.16} & & \makecell[c]{9.55} & \rd{43.34} & & \makecell[c]{0.19} & {0.17} & \rd{1.33} & & \makecell[c]{11.75}\\
       \hline

       LVI-SAM\cite{shan2021lvi}, 2021  & \makecell[c]{0.85} & \makecell[c]{136.03} & \makecell[c]{4.23} & \makecell[c]{30.85} & & \makecell[c]{7.06} & \nd \sl{28.44} & & \makecell[c]{0.64} & \rd{0.49} & \makecell[c]{7.63} & & \makecell[c]{12.56}\\
       
       R2LIVE\cite{lin2021r2live}, 2021 & \nd \sl 0.11 & \nd \sl{0.11} & \nd \sl{0.13} & \fs \bf{0.10} & & \blackx & \blackx & & \fs \bf{0.09} & \nd \sl{0.19} & \rd{1.33} & & \rd {1.36}\\
       
       R3LIVE\cite{lin2021r3live}, 2022  & \makecell[c]{8.76} & \makecell[c]{4.24} & \makecell[c]{1.12} & \makecell[c]{9.00} & & \rd {6.07} & \blackx & & \makecell[c]{1.07} & \makecell[c]{6.00} & \makecell[c]{1.69} & & \blackx\\
       
      Fast-LIVO\cite{zheng2022fast}, 2022  & \blackx & \makecell[c]{8.95} & \blackx  & \makecell[c]{9.49} & & \makecell[c]{7.96} & \blackx & & \makecell[c]{0.78} & \makecell[c]{1.92} & \makecell[c]{1.50} & & \blackx\\
      
       Coco-LIC\cite{lang2023coco}, 2023  & \makecell[c]{1.77} & \makecell[c]{0.97} & \makecell[c]{0.54} & \makecell[c]{1.66} & & \makecell[c]{6.98} & \blackx & & \makecell[c]{0.64} & \makecell[c]{1.80} & \nd \sl{1.21} & & \nd \sl{0.54}\\
       
       SR-LIVO\cite{yuan2024sr}, 2024  & \makecell[c]{1.23} & \makecell[c]{0.28} & \fs\bf{0.09} & \rd {1.31} & & \blackx & \blackx & & \makecell[c]{0.86} & \blackx & \fs \bf {0.09} & & \makecell[c]{92.14}\\

       Fast-LIVO2\cite{zheng2024fast}, 2024  & \rd{0.44} & \rd{0.28} & \rd{0.17} & \sl \nd{0.33} & & \nd \sl{3.35} & \blackx & & \rd{0.51} & \makecell{0.81} & \makecell[c]{9.71} & & \fs \bf{0.09}\\

       \bf{Ground-Fusion++(Ours)},2025  & \fs \bf{0.09} & \fs \bf{0.09} & \rd{0.17} & \fs \bf{0.10} & & \fs \bf {1.00} & \fs \bf{5.28} & & \nd \sl{0.10} & \fs \bf{0.09} & \rd{1.33} & & {5.20}\\

        \hline
    \end{tabular}
    \end{adjustbox}
    \vspace{-3mm}

\end{table*}

To systematically validate the effectiveness of the proposed M3DGR benchmark, we evaluate cutting-edge SLAM systems on ten sequences selected from M3DGR dataset. The characteristics of these sequences are summarized in Table~\ref{sample_tab}. Quantitative results comparing all baselines are presented in Table~\ref{rmse_tab}.
\vspace{-2mm}

\subsection{Baselines}
\vspace{-2mm}
we conducted comprehensive evaluations across three SLAM categories: vision-based approaches, LiDAR-based approaches, and LiDAR-vision fusion systems.
For vision-based methods, we assess pure visual odometry\cite{wang2021tartanvo} and visual-inertial odometry systems\cite{mur2017orb,campos2021orb,von2022dm,qin2018vins}, along with their extensions incorporating\cite{shan2019rgbd}, GNSS\cite{cao2022gvins}, or wheel odometry\cite{Tingda2022VIW,Wallong2021}. In particular, Ground-Fusion\cite{yin2024ground} is a tightly-coupled RGB-Depth-Wheel-IMU-GNSS system within a factor graph.
For LiDAR-based methods, we evaluate classical LiDAR SLAM\cite{qin2018aloam,lin2020loam,chengwei2023CTLO} and LIO approaches\cite{shan2018lego,li2021towards,xu2022fast,bai2022faster,he2023point,huang2023log,chengwei2023CT,Livox2021LIO,chen2023direct,qin2020lins,ma2024mm,chengwei2023hmlio,chengwei2022iESKFlio,yuan2022efficient,ye2019tightly}.
For LiDAR-visual SLAM, both tightly-coupled frameworks\cite{lin2021r2live,lang2023coco,zheng2022fast,zheng2024fast,lin2021r3live} and loosely-coupled\cite{shan2021lvi} are evaluated, along with our proposed system. Furthermore, Additionally, we evaluate wheel odometry integration and GNSS-based Single Point Positioning (SPP) solutions.

\vspace{-2mm}

\subsection{Benchmark Tests}

\vspace{-1mm}

\textbf{Overall Performance:} 
Most vision-based methods\cite{von2022dm,wang2021tartanvo,qin2018vins} perform well in small-scale environments with abundant visual features but struggle in challenging scenarios, often leading to tracking failures. In contrast, Ground-Fusion\cite{yin2024ground} achieves more stable localization, benefiting from multi-sensor fusion, though it still drifts sometimes. LiDAR-based SLAM generally offers higher localization accuracy, particularly in large-scale outdoor environments. Among the evaluated methods, CT-LIO\cite{chengwei2023CT}, HM-LIO\cite{chengwei2023hmlio} and IESKF-LIO\cite{chengwei2022iESKFlio} exhibit the most consistent and precise performance. Notably, continuous-time approaches\cite{chengwei2023CT,chengwei2023CTLO} yield smoother and more stable trajectories compared to discrete-time methods such as A-LOAM\cite{qin2018aloam} and LIO-SAM\cite{shan2020lio}. For LiDAR-visual SLAM, R2LIVE\cite{lin2021r2live}, Ground-Fusion++, and Fast-LIVO2\cite{zheng2024fast} exhibit the most reliable performance. While their localization accuracy offers only marginal improvements over leading LiDAR-based methods like\cite{chengwei2023hmlio, chengwei2023CT}, they significantly enhance mapping quality by leveraging RGB information, which enriched structural details.

\begin{figure}
    \centering
    \begin{tabular}{cc}
        \includegraphics[width=0.19\textwidth]{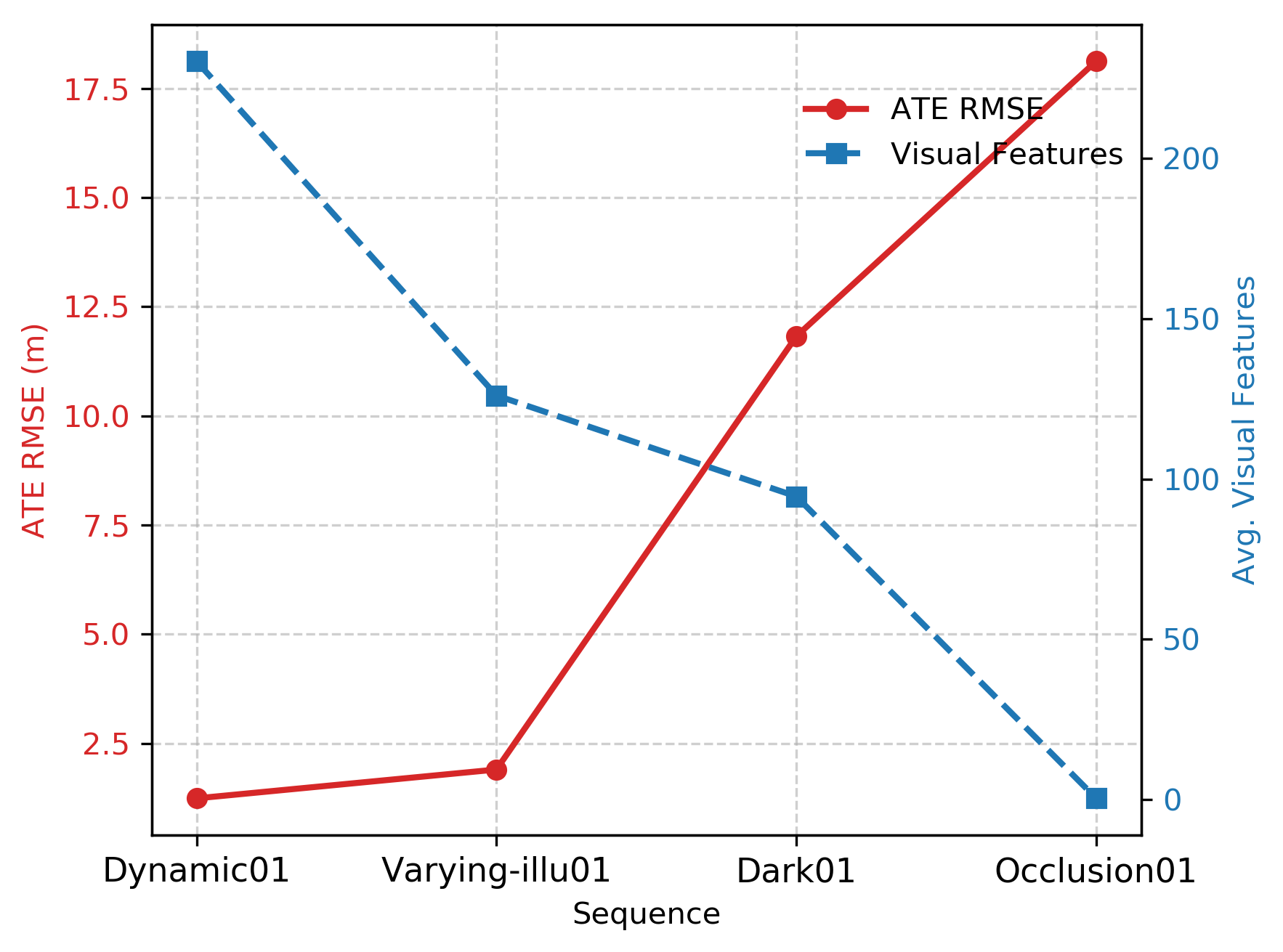}&
        \includegraphics[width=0.19\textwidth]{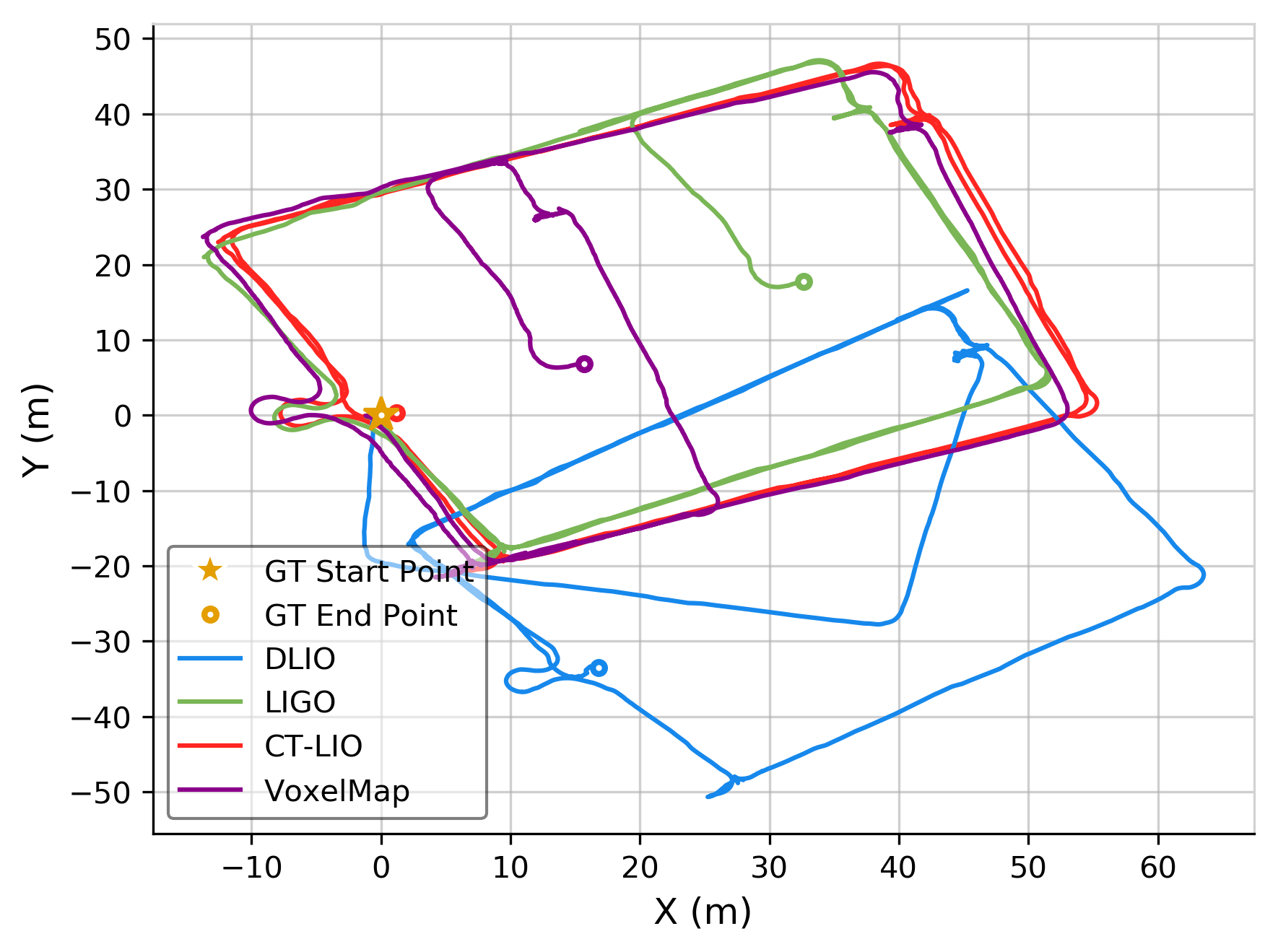} 
         \\
        (a) & (b) \\
        \includegraphics[width=0.19\textwidth]{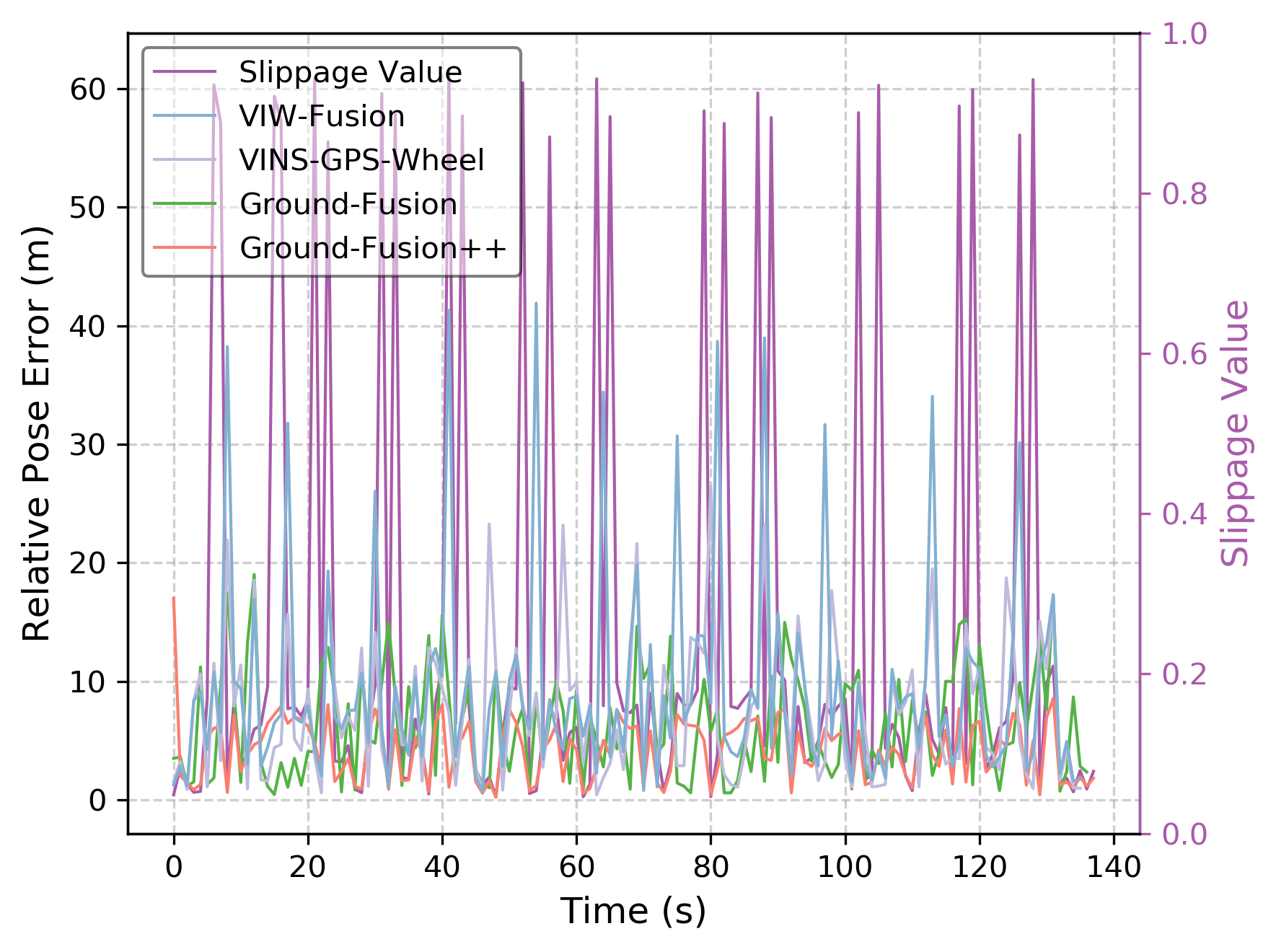}&
        \includegraphics[width=0.19\textwidth]{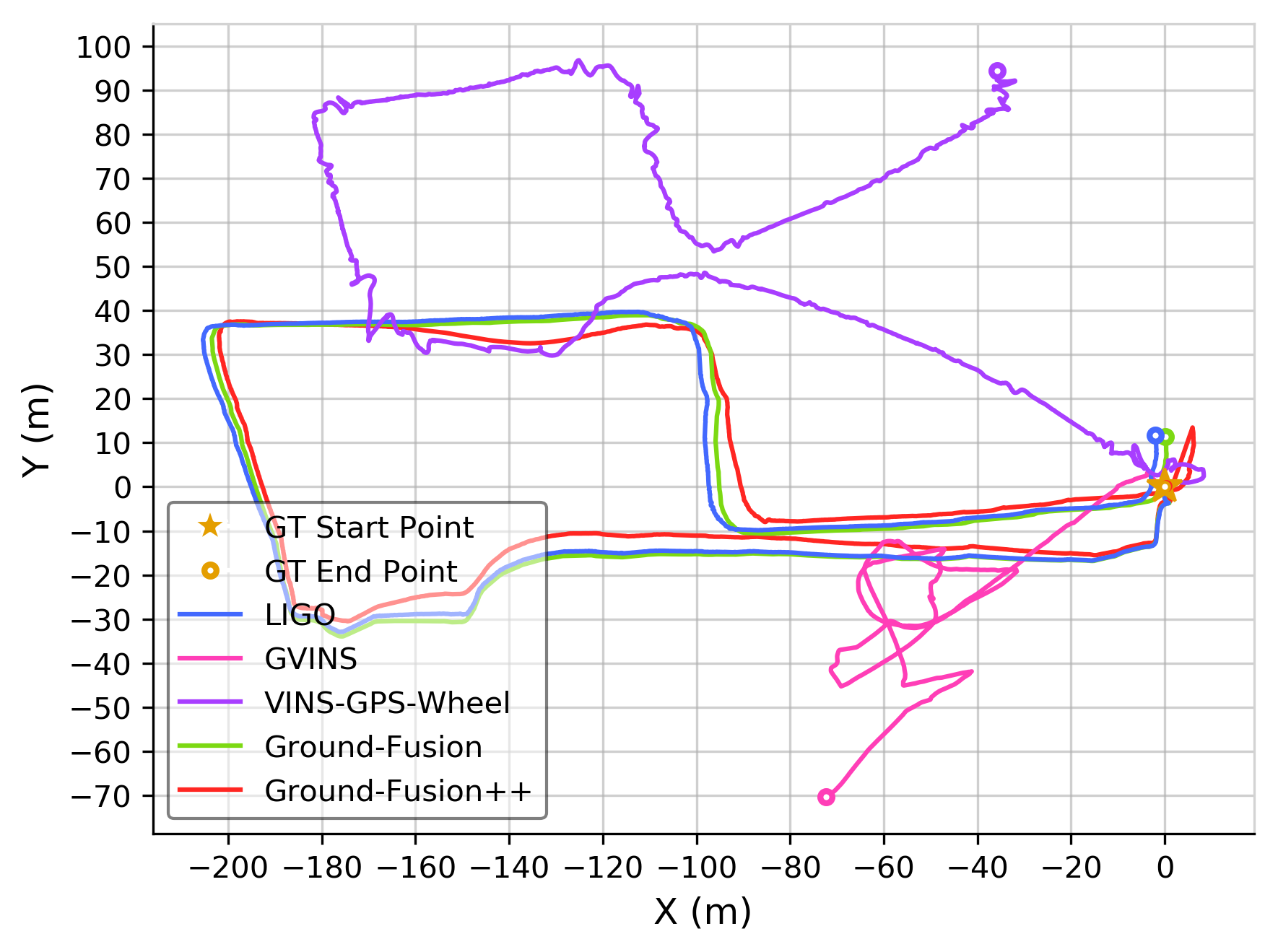} 
         \\
        (c) & (d) \\
    \end{tabular}
    \vspace{-3mm}
    \caption{(a) Avg. ATE RMSE and visual feature count across different challenging visual sequences. We set it as 20m if a system fails. (b) Trajectories of different LiDAR methods on the $Elevator01$ sequence. (c) Wheel slippage and RPE for different wheel-based methods. (d) Trajectories of different GNSS-integrated systems in the GNSS-denial01 sequence.}
    \label{fig:combined}
    \vspace{-5mm}
\end{figure}
\textbf{Visual Challenge:}
As the scene transitions from the dynamic environment to the varying lighting conditions, darkness, and eventually complete occlusion, the availability of reliable feature points decreases progressively. Consequently, the average ATE RMSE of all tested vision-based SLAM algorithms steadily increases, as illustrated in Figure~\ref{fig:combined} (a). Notably, systems incorporating wheel odometry\cite{Tingda2022VIW,Wallong2021,yin2024ground} demonstrate greater robustness in these visual challenges, demonstrating the effectiveness of multi-sensor fusion in improving SLAM performance.

\textbf{LiDAR Degeneracy:}
In highly structured corridors(the $Corridor01$ sequence), LiDAR-based algorithms such as\cite{xu2022fast,he2023point} frequently experience severe drift or even complete localization failure due to the lack of distinctive geometric features for reliable pose estimation. Certain systems, including VoxelMap\cite{yuan2022efficient}, HM-LIO\cite{chengwei2023hmlio}, and Fast-LIVO2\cite{zheng2024fast}, demonstrate better robustness in these cases. In the $Elevator01$ sequence, most systems fail when the elevator goes up. Only CT-LIO \cite{chengwei2023CT} maintains good localization, with temporary vertical drift during elevator movement while keeping stable horizontal positioning.

\textbf{Wheel Slippage:}
Most wheel odometry-integrated algorithms\cite{Tingda2022VIW,Wallong2021} suffer from significant drift when encountering wheel slippage. However, Ground-Fusion\cite{yin2024ground} mitigates this issue through a wheel anomaly detection mechanism that filters out unreliable observations, as shown in Figure~\ref{fig:combined} (c). This highlights that even a simple degradation-handling method can significantly enhance robustness.

\textbf{GNSS Denial:}
In GNSS-denied environments, most GNSS-integrated SLAM systems\cite{cao2022gvins} experience substantial performance degradation due to the lack of absolute positioning updates. However, systems such as VINS-GPS-Wheel\cite{Wallong2021} and Ground-Fusion\cite{yin2024ground}, which incorporate wheel odometry, exhibit reduced drift under these conditions by relying on additional motion constraints. Their trajectories on the GNSS-Denial01 sequence are presented in Figure~\ref{fig:combined}(d).

\begin{table}[h]
\caption{Comparison of localization accuracy (m) under LiDAR degradation scenarios with different configurations.}
\centering
\begin{adjustbox}{width=0.65\columnwidth}
\begin{tabular}{*{3}c}
\hline
\textbf{Methods} & \textbf{corridor01}$^\dagger$ & \textbf{corridor02}$^\dagger$ \\ \hline
LIO subsystem & 6.14 & 26.66 \\ 
Ground-Fusion++ with~\cite{ji2024point} & 4.72 & 2.12 \\ 
Ground-Fusion++ (Ours) & \fs \bf{1.67} & \fs \bf{1.73} \\ 
\hline
\end{tabular}
\end{adjustbox}
\label{tab:Switching_system}
\begin{flushleft}
\footnotesize{$^\dagger$ Artificial noise was injected during two degradation intervals: 50--80s and 200--250s.}
\end{flushleft}
\vspace{-6mm}
\end{table}

\subsection{Effectiveness of Our System}

\textbf{Degeneration-adaptive Capability:}
To validate the robustness of our system under LiDAR degradation, we conduct comparative experiments using three configurations: the LIO subsystem alone, Ground-Fusion++ integrated with the degeneration detection method proposed in \cite{ji2024point}, and our full Ground-Fusion++ system with the proposed degeneration detection and adaptive sensor fusion strategy. As shown in Table~\ref{tab:Switching_system}, our method achieves significantly better localization accuracy under harsh LiDAR degradation.

\textbf{Long-term Capability:}
We conduct experiments on the $Longtime01$ sequence, which spans a trajectory of about 30 minutes in a large-scale outdoor campus environment. Our proposed system, Ground-Fusion++, achieves an ATE RMSE of 7.5m with adaptive sensor strategy, significantly outperforming Ground-Fusion (22.5m) \cite{yin2024ground}. Moreover, while Ground-Fusion fails to generate a dense map in this challenging scenario, Ground-Fusion++ successfully constructs a high-quality, colorized dense map, as shown in Figure \ref{gf_immesh}(c). 


\begin{figure}[htbp]
    \centering
    \includegraphics[width=0.65\columnwidth]{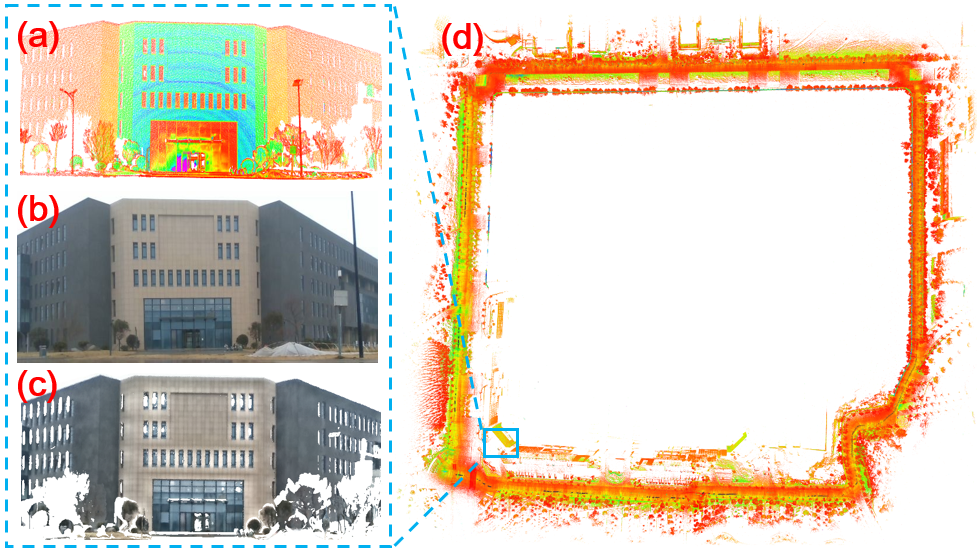}
    \caption{The reconstruction results of Ground-Fusion++ on the $Longtime01$ sequence. (a) The input RGB image. (b) The corresponding point cloud. (c) The dense color mesh generated by Ground-Fusion++. (d) The full point cloud.}
    \label{gf_immesh}
    \vspace{-3mm}
\end{figure}

\section{Conclusion}
This work presents M3DGR, a large-scale multi-sensor dataset for ground robot SLAM under challenging conditions. By systematically inducing sensor degradation scenarios, M3DGR enables in-depth evaluation of multi-sensor fusion algorithms. We evaluate 40 leading SLAM systems with diverse sensor settings across M3DGR, uncovering their limitations in extreme environments. 
To further support robust SLAM development, we propose Ground-Fusion++, a modular degradation-aware multi-sensor fusion framework with adaptive sensor selection, which demonstrates resilient localization and strong mapping ability. 




\small
\bibliographystyle{IEEEtrans}
\bibliography{root}

\end{document}